\documentclass[11pt]{article}
\usepackage{graphicx}
\usepackage{amsmath}
\usepackage{amsthm}
\usepackage{ulem}
\usepackage{color, soul}
\usepackage{float}
\usepackage{mathtools}
\usepackage{amsmath,amsfonts}
\usepackage{algorithmic}
\usepackage{algorithm}
\usepackage{array}
\usepackage[caption=false,font=normalsize,labelfont=sf,textfont=sf]{subfig}
\usepackage{textcomp}
\usepackage{stfloats}
\usepackage{url}
\usepackage{verbatim}
\usepackage{graphicx}
\usepackage{cite}
\usepackage[numbers]{natbib}
\usepackage{multirow}
\usepackage{color}

\newtheorem{theorem}{Theorem}

\newtheorem*{example*}{Example}

\newtheorem*{lemma*}{Lemma}

\DeclareMathOperator*{\argmax}{arg\,max}

\newcommand{\beq}{\begin{equation}}
\newcommand{\eeq}{\end{equation}}

\newcommand{\util}{\widetilde{u}}

\newcommand{\vtil}{\widetilde{v}}
\newcommand{\bvtil}{\widetilde{\bv}}

\newcommand{\bv}{\mathbf{v}}

\newcommand{\bx}{\mathbf{x}}
\newcommand{\by}{\mathbf{y}}
\newcommand{\bz}{\mathbf{z}}

\newcommand{\bbeta}{\boldsymbol{\beta}}

\newcommand{\btheta}{\boldsymbol{\theta}}

\begin{document}

\baselineskip=21pt

\begin{center}
{\Large \textbf{Deep Neural Networks for Semiparametric Frailty Models via H-likelihood}} \\[0pt]
\bigskip Hangbin Lee$^{1}$, Il Do Ha$^{2, *}$ and Youngjo Lee$^{1}$\\[0pt]
\bigskip {\small $^{1}$Department of Statistics, Seoul National University,
Seoul, South Korea\\[0pt]
$^{2}$Department of Statistics, Pukyong National University, Busan 48513, South
Korea\\[0pt]
$^{*}$Corresponding author. \textit{E-mail address}: idha1353@pknu.ac.kr
(I.D. Ha)}
\end{center}

\bigskip

\begin{abstract}
\noindent For prediction of clustered time-to-event data, we propose a new
deep neural network based gamma frailty model (DNN-FM). An advantage of the
proposed model is that the joint maximization of the new h-likelihood
provides maximum likelihood estimators for fixed parameters and best
unbiased predictors for random frailties. Thus, the proposed DNN-FM is trained by
using a negative profiled h-likelihood as a loss function, constructed by profiling out the non-parametric baseline hazard. Experimental studies
show that the proposed method enhances the prediction performance of the
existing methods.
A real data analysis shows that the inclusion of subject-specific frailties
helps to improve prediction of the DNN based Cox model (DNN-Cox).
\newline

\end{abstract}

\section{Introduction}

Recently, deep neural network (DNN) has provided a major breakthrough to
enhance prediction in various areas (LeCun et al., 2015; Goodfellow, 2016).
The DNN models allow extensions of Cox proportional hazards (PH) models
(Kvamme et al., 2019; Sun et al.,2020).
Recently, subject-specific prediction of the DNN models has been studied by
including random effects in neural network (NN) predictor (Tran et al.,
2020; Mandel et al., 2022). However, these DNN random-effect models have
been studied for only complete data. In this paper we propose a new
DNN-FM. To the best of our knowledge, there
is no literature on the DNN-FM for censored survival data. Lee and Nelder
(1996) introduced the h-likelihood for the inference of general models with
random effects and Ha, Lee and Song (2001) extended it to the
semi-parametric frailty models. We reformulate the h-likelihood to obtain
maximum likelihood estimators (MLEs) for fixed unknown parameters and best
unbiased predictors (BUPs; Searle et al., 1992; Lee et al., 2017) for
random frailties by a simple joint maximization of the profiled
h-likelihood, which is constructed by profiling out the non-parametric baseline hazard
for semi-parametric DNN-FMs. Thus, the proposed DNN-FM can be
trained by using a negative profiled h-likelihood as a loss function.
Experimental  studies show that the proposed method enhances the prediction
performance of the existing DNN-Cox and FM in terms of Brier
score and C-index, which are popular predictive measures in survival
analysis.

In Section 2, we review the DNN-Cox model. We propose the DNN-FM and
introduce its h-likelihood in Section 3 and learning
algorithm in Section 4. The experimental study is presented 
to compare its predictive performance with various methods in Section 5.  
A real data analysis is in Section 6, followed by concluding remarks in Section 7.
A theoretical framework for an online learning and all the technical details are in Appendix.

\section{A review of DNN-Cox model}

\subsection{DNN-Cox model}

Let $T_{i}$ be the survival time (time-to-event) for subject $i=1,\ldots ,n,$
and let $\boldsymbol{x}_{i}=(x_{i1},\ldots ,x_{ip})^{T}$ be a $p$%
-dimensional vector of input variables (covariates or features). The
semi-parametric Cox model is as follows: For a given $\boldsymbol{x}_{i},$
the hazard function of $T_{i}$ is
\begin{equation}
\lambda _{i}(t|\boldsymbol{x}_{i})=\lambda _{0}(t)\exp (\eta _{i}),~~\eta
_{i}=\boldsymbol{x}_{i}^{T}\boldsymbol{\beta}
\end{equation}
where $\lambda _{0}(\cdot )$ is a non-parametric baseline hazard function,
the linear predictor $\eta _{i}=\boldsymbol{x}_{i}^{T}\boldsymbol{\beta}$ is
a parametric model for risk function (or risk score) of covariates $\boldsymbol{x}
_{i}$, and $\boldsymbol{\beta}$ is a vector of $p$-dimensional regression
parameters without intercept (or bias) term.
The survival function for $T_{i}$ given $\boldsymbol{x_{i}}$ is
\begin{equation*}
S(t|\boldsymbol{x_{i}})= P(T_{i}>t|\boldsymbol{x_{i}}) = \exp \{-\Lambda _{0}(t)e^{\eta _{i}}\},
\end{equation*}
where $\Lambda _{0}(t)$ is the baseline cumulative hazard.

The Cox model (1) is extended to the DNN-Cox model, by relaxing the
parametric linear model $\eta _{i}=\boldsymbol{x}_{i}^{T}\boldsymbol{\beta}$
with a non-linear function of $\boldsymbol{x}_{i}$,
\begin{equation}
\eta _{i}=\sum_{k=1}^{p_{L}}\beta _{k} g_{k}^{(L)}(\boldsymbol{x}_{i};
\boldsymbol{w})\equiv NN(\boldsymbol{x}_{i};\boldsymbol{w},\boldsymbol{\beta}%
),
\end{equation}
where $NN(\cdot )$ denotes neural network risk predictor of the output layer with
the $L$th (last) hidden layer; $\boldsymbol{\beta}=(\beta _{1},\cdots
,\beta _{p_{L}})^{T}$ is a vector of the output weights, with $p_{L}$ number
of nodes of the $L$th hidden layer; and
$\boldsymbol{w}=(\boldsymbol{w}_{1}^{T},\boldsymbol{w}_{2}^{T},\ldots ,\boldsymbol{w}_{L}^{T})^{T}$
is a combined vectorization consisting of a vector $\boldsymbol{w}_{\ell}$  $(\ell =1,2,\ldots ,L)$ of the $\ell$th hidden weights.
Here, the $g_{k}^{(L)}(\boldsymbol{x}_{i}; \boldsymbol{w})$ is the $k$-th node of the last hidden layer $g^{(L)}(
\boldsymbol{x}_{i};\boldsymbol{w})=(g_{1}^{(L)T}(\boldsymbol{x}_{i};
\boldsymbol{w}),\ldots ,g_{p_{L}}^{(L)T}(\boldsymbol{x}_{i};\boldsymbol{w}
))^{T}$, which depends on the input variables $\boldsymbol{x}_{i}$ and the hidden weights $\boldsymbol{w}$,
and the last hidden layer can
be expressed as the form of compositional functions
\begin{equation*}
g^{(L)}(\boldsymbol{x}_{i};\boldsymbol{w})=\sigma ^{(L)}(\cdots \sigma
^{(2)}(\sigma ^{(1)}(\boldsymbol{x_{i}};\boldsymbol{w}_{1});\boldsymbol{w}%
_{2})\cdots ;\boldsymbol{w}_{L}),
\end{equation*}
where $\sigma ^{(\ell )}(\cdot )$ denotes the activation function of hidden
layer for each $\ell =1,2,\ldots ,L$, and each of the $\boldsymbol{w}_{\ell }$ vectors includes the bias term.

In survival analysis, the observable random variables are, for $i=1,\ldots
,n $
\begin{equation*}
y_{i}=\mathrm{min}(T_{i},C_{i})\text{ \ and \ }\delta _{i}=I(T_{i}\leq C_{i})
\end{equation*}
where $C_{i}$ is the censoring time corresponding to $T_{i}$. The DNN
weights $(\boldsymbol{w},\boldsymbol{\beta})$ in (2) can be estimated by
minimizing the negative Breslow (1972) log-likelihood (denoted by $-\ell $;
Kvamme et al., 2019; Tarkhan and Simon, 2022) as a loss function, given by
\begin{equation}
\ell =\sum_{i}\delta _{i}\eta _{i}-\sum_{k}d_{(k)}\log \biggl \{\sum_{~i\in
R_{(k)}~}\exp (\eta _{i})\biggl \},
\end{equation}
where $\eta _{i}=NN(\boldsymbol{x}_{i}; \boldsymbol{w},\boldsymbol{\beta})$
is the NN predictor which also represents an output node of
the DNN-Cox model, $R_{(k)}=\{i:y_{i}\geq y_{(k)}\}$ is the risk set at
time $y_{(k)}$ which is the $k$th $(k=1,\ldots ,K)$ smallest distinct event
time among the $y_{i}$'s, and $d_{(k)}$ is the number of events at $y_{(k)}$.
The weights $(\boldsymbol{w},\boldsymbol{\beta})$ are usually obtained by
optimizing the loss function based on the gradient decent method.

\subsection{Prediction measures}

For censored data, the two popular measures, namely the Brier score and the
concordance index (C-index), have been used to evaluate the predictive
performance of the DNN-Cox model (2) (Kvamme et al., 2019).

\subsubsection{Brier score}

The time-dependent Brier score is defined as
\begin{equation*}
BS(t)=E\left\{ I(t)-S(t|\boldsymbol{x})\right\} ^{2},
\end{equation*}
where $BS(t)$ is the mean squared error of the difference between $I(t)$ and $S(t|\boldsymbol{x})$.
Here, $I(t)$ is the event status at the time point $t$ (i.e. $I(t)=I(T>t)=1$ if $T>t$ and $0$ otherwise) 
and $S(t|\boldsymbol{x})$ is a model-based survival function.
The estimated
Brier score (Graf et al., 1999) is given by
\begin{equation*}
\widehat{BS}(t)=\frac{1}{n}\sum_{i=1}^{n}\hat{w_{i}}(t)\left\{ y_{i}(t)-%
\hat{S}(t|\boldsymbol{x}_{i})\right\} ^{2},
\end{equation*}
where $y_{i}(t)=I(y_{i}>t)$ at a specific time point $t$ and $\hat{S}(t|%
\boldsymbol{x}_{i})$ is estimated survival function given $\boldsymbol{x}_{i}
$. Here, $\hat{w}_{i}(t)$ is the inverse probability of censoring weights
(IPCW)
\begin{equation*}
\hat{w_{i}}(t)=\frac{(1-y_{i}(t))\delta _{i}}{\hat{G}(y_{i})}+\frac{y_{i}(t)%
}{\hat{G}(t)},\ \mathrm{with}\ \hat{G}(t)=\hat{P}(C>t),
\end{equation*}
and $\hat{G}(\cdot )$ indicates the estimated survival function of censoring
time. Thus, the estimated Brier score can be viewed as the mean squared
error between the observed event status $y_{i}(t)$ and the predicted
survival function $\hat{S}(t|\boldsymbol{x}_{i})$. The lower Brier score
indicates a better predictive performance. For predictive performance of
Brier score at all available times, the integrated Brier score (IBS) is
usually used with the maximum survival time $t_{max}$:
\begin{equation*}
\mathrm{IBS}=\frac{1}{t_{max}}\int_{0}^{t_{max}}\mathrm{BS}(s)ds.
\end{equation*}

\subsubsection{C-index}

The definition of C-index is based on the property that a survival model
should predict a shorter survival time for subjects that fail earlier and a
longer survival time for subjects that fail later. Let $T_{i}$ and $T_{j}$
be independent survival times with corresponding covariate vectors $%
\boldsymbol{x}_{i}$ and $\boldsymbol{x}_{j}$, respectively. Then the C-index
is defined by
\begin{equation*}
C=P({S}(t|\boldsymbol{x}_{i})>{S}(t|\boldsymbol{x}_{j})|T_{i}>T_{j})=P(\eta
_{i}<\eta _{j}|T_{i}>T_{j}).
\end{equation*}
where $\eta _{k}=NN(\boldsymbol{x}_{k};\boldsymbol{w},\boldsymbol{\beta})$
are the NN predictors of the DNN-Cox model (2). Following Harrell et al.
(1996), the C-index can be estimated by
\begin{equation*}
\widehat{C}=\frac{\sum_{i}\sum_{j}\delta _{i}I(y_{i}<y_{j})\{I(\hat{\eta}%
_{i}>\hat{\eta}_{j})+0.5I(\hat{\eta}_{i}=\hat{\eta}_{j})\}}{%
\sum_{i}\sum_{j}\delta _{i}I(y_{i}<y_{j})},
\end{equation*}
where $\hat{\eta}_{k}=NN(\boldsymbol{x}_{k};\boldsymbol{\hat w},%
\boldsymbol{\hat\beta)}$. The range of C-index is from 0 to 1, and a larger
value indicates a better performance.

\section{Proposed DNN for frailty model}

The FMs have been introduced for prediction of clustered survival time.
Consider a clustered survival dataset
\begin{equation*}
D_{N}=\{(y_{ij},\delta _{ij},\boldsymbol{x}_{ij}),i=1,\ldots ,n;j=1,\ldots
,n_{i}\},
\end{equation*}
where $y_{ij}=\mathrm{min}(T_{ij},C_{ij})$ is the $j$th observation of the $i$th
subject (or cluster), $T_{ij}$ and $C_{ij}$ are the corresponding survival
and censoring times, respectively, and $\delta _{ij}=I(T_{ij}\leq C_{ij})$
is censoring indicator, and $\boldsymbol{x}_{ij}=(x_{ij1},\ldots
,x_{ijp})^{T}$ is a vector of $p$ covariates corresponding to $T_{ij}$.
Here, $n$ is the number of clusters, $n_{i}$ is cluster size and $%
N=\sum_{i=1}^{n}n_{i}$ is the total sample size. The dependency among $T_{ij}
$'s can be modelled via a frailty in the hazard function. Let $u_{i}$ denote
the unobserved frailty of the $i$th cluster. Then, the semi-parametric FM is
as follows. The conditional hazard function of $T_{ij}$ given $u_{i}$ and $%
\boldsymbol{x}_{ij}$ takes the form of
\begin{equation}
\lambda _{ij}(t|u_{i},\boldsymbol{x}_{ij})=\lambda _{0}(t)\exp (%
\boldsymbol{x}_{ij}^{T}\boldsymbol{\beta})u_{i}=\lambda _{0}(t)\exp (\eta
_{ij}),~~\eta _{ij}=\boldsymbol{x}_{ij}^{T}\boldsymbol{\beta}+v_{i},
\end{equation}
where $\eta _{ij}$ is linear predictor and $v_{i}=\log u_{i}$.

\subsection{DNN-FM}

The FM (4) is extended to a new DNN-FM by replacing $\boldsymbol{x}_{ij}^{T}%
\boldsymbol{\beta}$ with
\begin{equation}
NN(\boldsymbol{x}_{ij};\boldsymbol{w},\boldsymbol{\beta})=\sum_{k=1}^{p_{L}}
\beta _{k} g_{k}^{(L)}(\boldsymbol{x}_{ij};\boldsymbol{w}).
\end{equation}
In this paper, for the frailty $u_{i}$, we use a gamma distribution with $%
E(u_{i})=1$ and var$(u_{i})=\alpha $, which is denoted by Gamma$(1/\alpha
,\alpha )$. Figure 1 shows the extension of Cox model to the DNN-FM. Figure
2 presents a schematic diagram of architecture of the DNN-FM, which is
constructed by allowing for output nodes $NN(\boldsymbol{x}_{ij};%
\boldsymbol{\hat w},\boldsymbol{\hat\beta)}$ and $\hat{u}_{i}$ from the two
separate input layers, namely input vector $\boldsymbol{x}_{ij}$ and one-hot
encoding vector of subjects $\boldsymbol{z}_{i}$, respectively. In the DNN-FM,
subject-specific prediction can be made by multiplying the risk predictor 
$\exp(NN(\boldsymbol{x}_{ij};\boldsymbol{\hat w},\boldsymbol{\hat \beta}))$ and
frailty predictors $\hat{u}_{i}$.

\subsection{Construction of h-likelihood}

In FMs (4), it is important to define the likelihood to obtain the exact
MLEs for fixed parameters and BUPs for random frailties. Let $\boldsymbol{y}^{\ast
}=(y,\delta )$ with $y=\mathrm{min}(T,C)$. 
Let $\boldsymbol{\psi}=(\boldsymbol{\theta}^{T},\lambda _{0}(\cdot ))^{T}$ with $\boldsymbol{\theta}
=(\boldsymbol{\beta}^{T},\alpha )^{T}$ be the vector of fixed parameters.
Under the conditional independence and non-informative censoring given $v_{i}
$, Ha et al. (2001) proposed the use of a h-likelihood
\begin{equation}
\ell _{e}(\boldsymbol{\psi},\mathbf{v};\mathbf{y}^{\ast },\mathbf{v}%
)=\sum_{i,j}\log f_{\boldsymbol{\psi}}(y_{ij},\delta
_{ij}|v_{i})+\sum_{i}\log f_{\boldsymbol{\psi}}(v_{i}),
\end{equation}
where
\begin{equation*}
\log f_{\boldsymbol{\psi}}(y_{ij},\delta _{ij}|v_{i})=\delta _{ij}\{\log
\lambda _{ij}(y_{ij}|v_{i})\}-\Lambda _{ij}(y_{ij}|v_{i})=\delta _{ij}\{\log
\lambda _{0}(y_{ij})+\eta _{ij}\}-\Lambda _{0}(y_{ij})\exp (\eta _{ij})
\end{equation*}
is the conditional censored log-likelihood of $y_{ij}$ and $\delta _{ij}$
given $v_{i}$, $\Lambda _{0}(\cdot )$ is the cumulative baseline hazard
function, $f_{\boldsymbol{\psi}}(v_{i})$ is a density function of $v_{i}$
with the parameter $\boldsymbol{\psi}$, and $\mathbf{v}=\log (\mathbf{u})$.
Lee and Nelder's (1996) original aim of h-likelihood is to obtain MLEs for
all fixed parameters and good predictors for random effects by its joint
maximization. However, its joint maximization of the current h-likelihood
above cannot give an exact MLE for variance component $\alpha $. In this
paper, we introduce a new h-likelihood for the gamma FM (4). Consider an
extended  likelihood (Lee et al., 2017) in the $\mathbf{v}^{c}$ scale
\begin{equation}
h(\boldsymbol{\psi},\mathbf{v}^{c})=\ell (\boldsymbol{\psi};\mathbf{y}^{\ast
})+\log f_{\boldsymbol{\psi}}(\mathbf{v}^{c}|\mathbf{y}^{\ast }),
\end{equation}
where $\ell (\boldsymbol{\psi};\mathbf{y}^{\ast })=\log \int f_{%
\boldsymbol{\psi}}(\mathbf{y}^{\ast },\mathbf{v})d\mathbf{v}$ is the
marginal log-likelihood. Given $\boldsymbol{\psi}$, let $\tilde{\mathbf{v}}%
^{c}$ be
\begin{equation*}
\tilde{\mathbf{v}}^{c}=\arg \max_{\mathbf{v}^{c}}h(\boldsymbol{\psi},\mathbf{%
v}^{c})=\arg \max_{\mathbf{v}^{c}}f_{\boldsymbol{\psi}}(\mathbf{v}^{c}|%
\mathbf{y}^{\ast })
\end{equation*}
From (7), a sufficient condition for $h(\boldsymbol{\psi},\mathbf{v}^{c})$
to give the exact MLEs for $\boldsymbol{\psi}$ is that $f_{\boldsymbol{\psi}}(\tilde{%
\mathbf{v}}^{c}|\mathbf{y}^{\ast })$ is independent of $\boldsymbol{\psi}$.
Let
\begin{equation}
v_{i}^{c}=v_{i}\exp \left\{ a_{i}(\alpha ,\delta _{i+})\right\} ,
\end{equation}
where $a_{i}(\alpha ,\delta _{i+})=\left( \delta _{i+}+\alpha ^{-1}\right)
\left( \log (\delta _{i+}+\alpha ^{-1})-1\right) -\log \Gamma (\delta
_{i+}+\alpha ^{-1}).$ Appendix A.1 shows that the predictive likelihood
\begin{equation*}
\log f(\tilde{\mathbf{v}}^{c}|\mathbf{y}^{\ast })=\sum_{i=1}^{n}\log f_{%
\boldsymbol{\psi}}(\tilde{v}_{i}^{c}|\mathbf{y}^{\ast
})=\sum_{i=1}^{n}\left\{ \log f_{\boldsymbol{\psi}}(\tilde{v}_{i}|\mathbf{y}%
^{\ast })-a_{i}(\alpha ,\delta _{i+})\right\} =0
\end{equation*}
is free from $\boldsymbol{\psi}$. Thus, $\ell (\boldsymbol{\psi},\mathbf{y}%
^{\ast })=h(\boldsymbol{\psi},\tilde{\mathbf{v}}^{c})$. Let $h(%
\boldsymbol{\psi},\mathbf{v})$ be a reparameterization of the h-likelihood
(7). Then
\begin{equation*}
h=h(\boldsymbol{\psi},\mathbf{v})=\ell _{e}(\boldsymbol{\psi},\mathbf{v};%
\mathbf{y}^{\ast },\mathbf{v})+\log \left| \frac{d\mathbf{v}}{d\mathbf{v}^{c}
}\right| =h(\boldsymbol{\psi},\mathbf{v}^{c}),
\end{equation*}
where $\ell _{e}(\boldsymbol{\psi},\mathbf{v};\mathbf{y}^{\ast },\mathbf{v})$
is the h-likelihood (6) of Ha et al. (2001) and $\log \left| \frac{d\mathbf{v}}{d\mathbf{v}^{c}
}\right|=-\sum_{i=1}^{n} a_{i}(\alpha ,\delta _{i+})$. Thus, $h(\boldsymbol{\psi},%
\mathbf{v})\neq \ell _{e}(\boldsymbol{\psi},\mathbf{v};\mathbf{y}^{\ast },%
\mathbf{v})$. Given $\boldsymbol{\psi}$, we have the
BUP for $\mathbf{u}$, $E(\mathbf{u}|\mathbf{y^{\ast }})$,
by solving $\partial h/\partial \mathbf{v}=0$ (or $\partial
h/\partial \mathbf{u}=0)$, where $\mathbf{u}=\exp (\mathbf{v})$ (see Appendix A.1).
The joint maximization of the new
h-likelihood gives MLEs for whole parameters including variance component
and BUPs for random frailties.

\section{Learning algorithm using the profiled h-likelihood}

In the DNN-FM (5), the new h-likelihood is
\begin{align}
h=h(\boldsymbol{\psi},\mathbf{v})=& \sum_{i=1}^{n}\sum_{j=1}^{n_{i}}\Big[%
\delta _{ij}\{\log \lambda _{0}(y_{ij})+\eta _{ij}\}-\Lambda
_{0}(y_{ij})\exp (\eta _{ij})\Big]  \notag \\
& +\sum_{i=1}^{n}\left[ \frac{\log u_{i}-u_{i}}{\alpha }-\alpha ^{-1}\log
\alpha -\log \Gamma (\alpha ^{-1})-a_{i}(\alpha ,\delta _{i+})\right]
\end{align}
where
\begin{equation*}
\eta _{ij}=NN(\boldsymbol{x}_{ij};\boldsymbol{w},\boldsymbol{\beta})+v_{i}
\end{equation*}
and $a_{i}(\alpha ,\delta _{i+})$ is given in (8). For eliminating
the non-parametric baseline hazard $\lambda _{0}(\cdot )$ in (9), following Ha et al.
(2001) we can have a profiled h-likelihood,
\begin{equation}
h_{p}=h_{p}(\boldsymbol{\theta},\mathbf{v})=h_{\mathrm{PL}
}-\sum_{i=1}^{n}a_{i}(\alpha ,\delta _{i+}),
\end{equation}
where
\begin{equation*}
h_{\mathrm{PL}}=\sum_{ij}\delta _{ij}\eta _{ij}-\sum_{k}d_{(k)}\log \left[
\sum_{~(i,j)\in R_{(k)}~}\exp (\eta _{ij})\right] +\sum_{i=1}^{n}\left[
\frac{\log u_{i}-u_{i}}{\alpha }-\frac{\log \alpha }{\alpha }-\log \Gamma
\left( \frac{1}{\alpha }\right) \right]
\end{equation*}
is the penalized partial likelihood (PPL; Ripatti and Palmgren, 2000;
Therneau and Grambsch, 2000). Here, $R_{(k)}=\{(i,j):y_{ij}\geq y_{(k)}\}$
is risk set at time $y_{(k)}$ $(k=1,\ldots ,K)$, and $d_{(k)}$ is the number
of events at $y_{(k)}$ which is the $k$th $(k=1,\ldots ,K)$ smallest
distinct event times among the $y_{ij}$'s. Direct maximization of the PPL
cannot provide MLEs. To obtain the MLEs, Gu et al. (2014) proposed the use
of the marginal partial log-likelihood
\begin{equation*}
\ell _{p}=\log \int \exp (h_{\mathrm{PL}})dv.
\end{equation*}
However, this integration is often numerically intractable. Thus, Ha et al.
(2001, 2017) and Ripatti and Palmgren (2000) proposed the use of the Laplace
approximation of $\ell _{p}$, which is still numerically difficult and does
not give the exact MLEs. The Laplace approximation can yield a biased estimation for frailty models with a small cluster size or under heavy censoring (Jeon et al., 2012; Gorfine and Zuker, 2023).

An advantage of the h-likelihood approach is that the nuisance parameters
associated with the non-parametric hazard $\lambda _{0}(\cdot)$ can be
eliminated by profiling.
Since the joint maximization of $h_{p}$ gives the MLEs for fixed parameters and BUPs for random frailties, the DNN-FM (5) can be trained by using negative profiled
h-likelihood (i.e. $-h_{p}$) as a loss function, which contains $NN(\boldsymbol{x};\boldsymbol{w},\boldsymbol{\beta})$ and $u_{i}$. Thus, the two
separate output nodes are necessary as in Figure 2.

\subsection{Local minima problem}

In FMs, for identifiability we impose the constraints $\text{E}(u_{i})=1$
because for any $\epsilon$,
\begin{align*}
\lambda _{ij}(t|u_{i},\mathbf{x}_{ij})& =\lambda _{0}(t)\exp \left\{ \text{NN
}\left( \mathbf{x}_{ij};\mathbf{w},\boldsymbol{\beta}\right) \right\} u_{i}
\\
& =\lambda _{0}(t)\exp \left\{ \text{NN}\left( \mathbf{x}_{ij};\mathbf{w},%
\boldsymbol{\beta}\right) +\epsilon \right\} (u_{i}/\exp (\epsilon )).
\end{align*}
However, DNN models often encounter local minima which violates the constraints. This
causes a computational difficulty in the DNN-FM. To prevent poor prediction
due to the local minima, we introduce an adjustment on the predictor of $u_{i}$
\begin{equation}
\widehat{u}_{i}\leftarrow \frac{\widehat{u}_{i}}{\frac{1}{n}\sum_{i=1}^{n}%
\widehat{u}_{i}}
\end{equation}
to satisfy
\begin{equation*}
\frac{1}{n}\sum_{i=1}^{n}\widehat{u}_{i}=1.
\end{equation*}

\subsection{ML learning algorithm}

We propose a h-likelihood learning algorithm:

\begin{itemize}
\item  \textbf{Inner loop:} For given $\widehat{\alpha }$, find optimal $(%
\widehat{\mathbf{w}},\widehat{\boldsymbol{\beta}},\widehat{\mathbf{u}})$
under a loss function $-h_{p}$ (10).

\item  \textbf{Adjustment:} Transport $\widehat{\mathbf{u}}$ as in (11).

\item  \textbf{Outer loop:} For given $(\widehat{\mathbf{w}},\widehat{%
\boldsymbol{\beta}},\widehat{\mathbf{u}})$, find an optimal $\widehat{\alpha
}$ under a loss function $-h_{p}$ (10).
\end{itemize}

This algorithm describes double loop iterative procedures with an additional
adjustment on the frailty predictors: for details, see \textbf{Algorithm 1}.
Figure 3 displays a schematic diagram of the h-likelihood learning procedure of
the DNN-FM (5).

\begin{algorithm}[H]
\caption{H-likelihood Learning Algorithm.} \label{al}
\begin{algorithmic}
\STATE
\STATE \hspace{-0.5cm} {\textsc{Repeat until $\alpha$ converges:}}
\STATE
\STATE {\textsc{Train the network:}}
\STATE \hspace{0.5cm}
	$\widehat{\mathbf{w}}, \widehat{\boldsymbol{\beta}}, \widehat{\mathbf{v}}
	\gets \underset{\boldsymbol{{\mathbf{w}, \boldsymbol{\beta}, \mathbf{v}}}}{\arg\min} \left\{ - h_{p}(\mathbf{w}, \boldsymbol{\beta}, \widehat{\alpha}, \mathbf{v}) \right\}$

\STATE \hspace{0.5cm}\textbf{return} $\widehat{\mathbf{w}}, \widehat{\boldsymbol{\beta}}, \widehat{\mathbf{v}}$
\STATE
\STATE {\textsc{Adjust the frailties:}}
\STATE \hspace{0.5cm}$\bar{u} \gets \sum_{i=1}^n \exp(\widehat{v}_i) / n$
\STATE \hspace{0.5cm}$\widehat{v}_i \gets \widehat{v}_i - \log \bar{u}$ \textbf{ for } $ i = 1,...,n $
\STATE \hspace{0.5cm}\textbf{return} $\widehat{\mathbf{v}}$
\STATE
\STATE {\textsc{Compute variance component:}}
\STATE \hspace{0.5cm}$
	\widehat{\alpha} 	\gets \underset{\alpha} {\arg\min}
	\left\{ - h_{p}(\widehat{\mathbf{w}},  \widehat{\boldsymbol{\beta}}, \alpha, \widehat{\mathbf{v}}) \right\}$
\STATE \hspace{0.5cm}\textbf{return} $\widehat{\alpha}$
\STATE
\end{algorithmic}
\label{alg1}
\end{algorithm}

\section{ Experimental studies}

To evaluate the performance of the proposed method, experimental studies are
conducted based on 100 replications of simulated data.
As performance measures, we use the extended forms of the IBS and
C-index of FMs (Oirbeek and Lesaffre, 2010, 2016) in Appendix A.2.

\subsection{Experimental design}

Given $u_{i}$ and $\mathbf{x}_{ij}$, survival times $T_{ij}$ are generated from the hazard function
\begin{equation*}
\lambda _{ij}(t|u_{i},\mathbf{x}_{ij})=\lambda _{0}(t)\exp \{f(\mathbf{x}
_{ij})\}u_{i},
\end{equation*}
where $f(\mathbf{x}_{ij})$ is an unknown true risk function of $x_{ij}$ and $%
\lambda _{0}(t)=\phi t^{\phi -1}$ is set to be a Weibull baseline hazard
with shape parameter $\phi =2$. The DNN-FM fits the true but unknown $f(%
\mathbf{x}_{ij})$ by $NN(\boldsymbol{x_{ij}};\boldsymbol{w},%
\boldsymbol{\beta})$. Here, the five input variables $\mathbf{x}%
_{ij}=(x_{1ij},...,x_{5ij})^{T}$ are generated from AR(1) process with
autocorrelation $\rho =0.5$ and frailties $u_{i}$ are generated from gamma
distribution with $\text{E}(u_{i})=1$ and $\text{Var}(u_{i})=\alpha $, and
\begin{align*}
f(\mathbf{x}_{ij})=& 0.4\cos (x_{1ij})+0.3\cos (x_{2ij})+0.6\cos
(x_{3ij})+0.5x_{2ij}\ast x_{3ij} \\
& +0.4/(x_{4ij}^{2}+1)+0.5/(x_{5ij}^{2}+1).
\end{align*}
We set the frailty variance $\alpha =0,0.5,1$ and $2$, where $\alpha =0$
means the DNN-Cox model without frailty. The censoring times are generated
from an exponential distribution with parameter empirically determined to
achieve approximately two right censoring rates, low (around 15\%) and high
(around 45\%). We set the total sample size $N=8000$ with $(n,n_{i})=(1000,8)
$ for all $i$. Thus, the dataset contains 1000 subjects and each subject has
8 observations. For each subject $i$, we assign 4 observations $(j=1,2,3,4)$
to the training set, 2 observations $(j=5,6)$ to the validation set and the
remaining 2 observations $(j=7,8)$ to the test set.

For comparison, we consider the fitting of the following four models.

\begin{itemize}
\item  \textbf{Cox:} Cox model (1)

\item  \textbf{DNN:} DNN-Cox model (2)

\item  \textbf{FM:} gamma frailty model (4)

\item  \textbf{DNN-FM:} The proposed DNN-FM (5)
\end{itemize}

The network architecture and hyper-parameters are tuned by using the vanilla
DNN-Cox. As an optimal result, we set all the DNN models to have 3 hidden
layers of 10 nodes with ``relu'' activation function. We use the full batch
and ``AdamW'' optimizer with learning rate 0.01. Early stopping with
validation loss is employed to prevent overfitting. The Cox model is
implemented using \texttt{lifelines} package in Python, gamma FM is
implemented using \texttt{frailtyEM} (Balan and Putter, 2019) package in R, and the DNN models
(DNN-Cox, DNN-FM) are implemented using Python-based Keras and Tensorflow.

\subsection{ Experimental results}

For evaluation of the prediction performances, IBS (12) and C-index (13) in Appendix 
are computed on the test set. Figure 4 shows box plots of IBS for each model under 15\% censoring. Figure 4(a) shows that all models have
comparable results when there is no frailty. Even if there is no frailty $%
(\alpha =0)$, the proposed DNN-FM model is still comparable to the vanilla
DNN-Cox, which should have the smallest IBC. Figure 4(b), (c) and (d) show
that the DNN-FM has the smallest IBS values when frailty is presented.
Figure 5 shows box plots of C-index for each model under 15\% censoring.
When there is no frailty term, Figure 5(a) shows that the two DNN models
(DNN-Cox, DNN-FM) have comparable results, but that the two non-DNN models
(Cox, FM) give poor results, which means they do not capture the nonlinear
effect of input variables in terms of C-index. As expected, Figure 5(b), (c)
and (d) show that the DNN-FM has the highest C-index. Next, Figures
6 and 7 present box plots of IBS and C-index under 45\% censoring,
respectively and they overall show similar trends to Figures 4 and 5.
However, the trends in Figure 6(a) are somewhat different. That is, the two
standard models (Cox and FM) in Figure 6(a) give poor results as compared to
those in Figure 4(a), meaning that under 45\% censoring, they do not again
capture the nonlinear effect of input variables in terms of IBC.

Mean and standard deviation of IBS and C-index for each model with two
censoring rates are summarized in Table 1. This confirms that the DNN-FM
outperforms three existing models (Cox, DNN-Cox and FM). Table 2 reports
mean and standard deviation of estimated frailty variance ($\hat{\alpha}$)
from train sets under 100 replications of simulated data. When $\alpha =0$,
the true model does not have frailties, and the estimates of $\alpha $ under
FM and DNN-FM with two censoring rates (15\% and 45\%) are closed to zero.
As $\alpha $ increases, the MLE of $\alpha $ under FM is downward biased,
whereas that under DNN-FM is consistent. As expected, we see that the
standard deviations of $\hat{\alpha}$ tend to increase as $\alpha $ or
censoring rate increases.

\section{Real data analysis: multi-center bladder cancer data}

We illustrate the DNN-FM method using a bladder cancer multi-center trial
conducted by the EORTC (Sylvester et al., 2006). We consider 392
bladder-cancer patients from 21 centers that participated in EORTC trial
30791.
The primary endpoint (event of interest) was time (day) to the first bladder
cancer recurrence from randomization. Of the 392 patients, 200 (51.02\%) had
recurrence of bladder cancer (event of interest) and 81 (20.66\%) died prior
to recurrence (a competing event). 111 (28.32\%) patients who were still
alive and without recurrence were censored at the date of the last available
follow-up.
Following Park and Ha (2019), we regarded the 81 competing risk events as
censored, resulting that censoring rate is 49.98\% with 192 censored
patients. The data are unbalanced due to different number of patients in
each center. In this paper, we used the data with 373 patients from 16 centers which have
more than 5 patients in each center. The numbers of patients per center
varied from 6 to 78, with mean 23.3 and median 17.5. In each center, we used
two randomly selected patients as test set, another two randomly selected
patients as validation set, and the remaining patients as training set.

We consider the following 12 categorical input variables ($\boldsymbol{x}$):

\begin{itemize}
\item  Chemotherapy as the main covariate (CHEMO; no=0, yes=1),

\item  Age (0 if Age$\leq $65 years, 1 if Age$>65$ years),

\item  Sex (male=0, female=1),

\item  Prior recurrent rate (PRIORREC; primary, $\leq $ 1/yr, $>$ 1/yr);%
\newline
PRIORREC1 = I(PRIORREC $\leq $ 1/yr), PRIORREC2 = I(PRIORREC $>$ 1/yr)

\item  Number of tumors (NOTUM; single, 2-7 tumors, $\geq $ 8 tumors);%
\newline
NOTUM1=I(NOTUM = 2-7 tumors), NOTUM2=I(NOTUM $\geq 8$ tumors);

\item  Tumor size (TUM3CM; 0 if Tumor size $<$3cm, 1 if Tumor size $\geq $%
3cm),

\item  T category (TLOCC; Ta=0, T1=1),

\item  Carcinoma in situ (CIS; no=0, yes=1),

\item  G grade (GLOCAL; G1, G2, G3);\newline
GLOCAL1=I(GLOCAL=G2), GLOCAL2=I(GLOCAL=G3).
\end{itemize}

Table 3 presents IBS and C-index on the test set of the bladder cancer data.
The DNN-FM shows the smallest IBS and the highest C-index which indicate the
best prediction performance, and DNN-Cox outperforms the two non-DNN models
(Cox and FM). In the train set, the estimated frailty variances are small
with $\hat{\alpha}=0.069$ for DNN-FM and $\hat{\alpha}=0.086$ for FM,
leading to similar values of IBS and C-index from the Cox and FM. Thus, in
this dataset the nonlinear effect of input variables are important in
predicting the survival probability of patients in each center. Figure 8
shows the time-dependent Brier scores on the test set under the four
models. Here, the Brier scores of the four models are similar at almost time
points before 3 years. However, after 3 years, the Brier scores of the proposed
DNN-FM are always noticeably lower than other three models. Accordingly, the
DNN-FM improves prediction of the DNN-Cox model.

\section{Concluding remarks}

We have presented a new DNN-FM. The joint maximization of its profiled
h-likelihood provides MLEs for fixed parameters and BUPs for random frailties. Our
empirical results demonstrate that the proposed method improves the
prediction performance of the existing DNN-Cox and FMs in terms of IBS and
C-index. The specification of the gamma frailty distribution in semi-parametric FMs
is insensitive to the estimates of fixed regression parameters (i.e.
weights) if the variance of frailty is not very large (Hsu et al. 2007; Ha
et al., 2001, 2017; Gorfine et al., 2023).
Extension of the proposed method
to other frailty distribution such as parametric (e.g. log-normal) or
non-parametric distribution (Chee et al., 2021) would be an interesting
further work.
The proposed DNN-FM can be trained for very large
clustered survival data by using an online learning, whose theoretical framework is in Appendix A.3.

\bigskip
\section*{Acknowledgements}
The research of Il Do Ha was supported by the National Research Foundation of Korea(NRF) grant funded by the Korea government(MSIT) (No. RS-2023-00240794). The research of Youngjo Lee
was supported by the National Research Foundation of Korea (NRF) grant funded by the Korea government (MSIT) (No.
2019R1A2C1002408).

\bigskip

\section*{Appendix: Supplementary Materials}

\subsection*{A.1 Derivation for the predictive likelihood}

Recall that $\mathbf{y}^{*}=(\mathbf{y}, \boldsymbol{\delta})$, where the $(i,j)$th component of $\mathbf{y}$ is
$y_{ij} =\mathrm{min}(T_{ij},C_{ij})$. Note that $\tilde{\mathbf{v}}$ is given by
\begin{align*}
\tilde{\mathbf{v}} &= \argmax_{\mathbf{v}} \left\{ \log f_{\boldsymbol{\psi}%
}(\mathbf{v}|\mathbf{y}^{*}) \right\} \\
&= \argmax_{\mathbf{v}} \left\{ \log f_{\boldsymbol{\psi}}(\mathbf{y}^{*}|%
\mathbf{v}) + \log f_{\boldsymbol{\psi}}(\mathbf{v}) - \log f_{%
\boldsymbol{\psi}}(\mathbf{y}^{*}) \right\} \\
&= \argmax_{\mathbf{v}} \left\{ \sum_{i=1}^n \sum_{j=1}^{n_i} \left(
\delta_{ij} v_i - \Lambda_{ij}^{(m)} e^{v_i} \right) + \sum_{i=1}^n \left(
\frac{ v_i - e^{v_i}}{\alpha} \right) \right\} \\
&= \argmax_{\mathbf{v}} \sum_{i=1}^n \left\{ v_i \left( \delta_{i+} +
\alpha^{-1} \right) - e^{v_i} \left( \Lambda_{i+} +\alpha^{-1} \right)
\right\} \\
&=\log \left( \frac{\delta_{i+} +\alpha^{-1} }{\Lambda_{i+} +\alpha^{-1} }
\right)
\end{align*}
where $\delta_{i+} = \sum_{j=1}^{n_i} \delta_{ij}$, $\Lambda_{i+} =
\sum_{j=1}^{n_i} \Lambda^{(m)}_{ij}=\sum_{j=1}^{n_i}\Lambda_{0}
(y_{ij})\exp(f(\boldsymbol{x}_{ij}))$. This implies that
\begin{equation*}
\tilde{u}_i = \exp(\tilde{v}_i)= \frac{\delta_{i+} +\alpha^{-1} }{%
\Lambda_{i+} +\alpha^{-1} } =E(u_{i}|\mathbf{y}_i^{*})
\end{equation*}
is the BUP (Searle et al., 1992) for $u_{i}(=\exp(v_{i}))$ in sense that it
gives minimum mean squared error of prediction (``best") and $E(\tilde{u}_i
-u_{i})=0$ (``unbiased") with $E(\tilde{u}_i)=E(u_{i})=1$, since
\begin{equation*}
u_i | \mathbf{y}_i^{*} \sim \text{Gamma} \left( \delta_{i+}+\alpha^{-1},
(\Lambda_{i+}+\alpha^{-1})^{-1} \right),
\end{equation*}
which is easily derived from the fact that the gamma distribution is
conjugate of the frailty model.
From the density function of gamma distribution above, the predictive
likelihood at $\tilde{\mathbf{v}}^c$ is given by
\begin{align*}
\log f(\tilde{\mathbf{v}}^c | \mathbf{y}^{*}) &= \sum_{i=1}^n \log f_{%
\boldsymbol{\psi}}(\tilde{v}^c_i | \mathbf{y}^{*}) = \sum_{i=1}^n \left\{
\log f_{\boldsymbol{\psi}}(\tilde{v}_i | \mathbf{y}^{*}) - a_i(\alpha,
\delta_{i+}) \right\} \\
&= \sum_{i=1}^n \left\{ \log f_{\boldsymbol{\psi}}(\tilde{u}_i | \mathbf{y}%
^{*}) + \log \tilde{u}_i - a_i(\alpha, \delta_{i+}) \right\} = 0,
\end{align*}
where $\tilde{u}_i = \exp(\tilde{v}_i)$.

\subsection*{A.2 Evaluation measures for DNN-FM}

The BS in Section 2.1 can be extended to the DNN-FM (5) as a conditional
form:
\begin{equation*}
BS_{c}(t)=E\left\{Y(t)-S(t|u,x)\right\}^2,
\end{equation*}
where $S(t|u,x)$ is the conditional survival function given $u$, and the
estimated conditional BS (Oirbeek and Lesaffre, 2016) is given by
\begin{equation}
\widehat{BS}_{c}(t)=\frac{1}{N} \sum_{ij \in D_N}\hat{w}_{ij}(t)\left%
\{y_{ij}(t)-\hat{S}(t|\hat u_{i}, x_{ij})\right\}^2,
\end{equation}
where $N=\sum_{i=1}^{N}$ is the total sample size and the IPCW is
\begin{equation*}
\hat{w}_{ij}(t) = \frac{(1-y_{ij}(t))\delta_{ij}}{\hat{G}(y_{ij})} + \frac{%
y_{ij}(t)}{\hat{G}(t)},\ \mathrm{with}\ \hat{G}(t)=\hat{P}(C>t).
\end{equation*}
The BS can be also summarized as the integrated Brier score (IBS) in Section
2.2.1.

The C-index in Section 2.2 can be also extended to the DNN-FM with clustered
survival data (Oirbeek and Lesaffre, 2010; Mauguen et al., 2013). For the
clustered data, we consider the overall conditional C-index, i.e. the
concordant probability defined for all comparable pairs; it can distinguish
two different types of pairs, within-cluster pairs and between-cluster
pairs, i.e. pairs whose members belong to the same cluster or to different
clusters, respectively. Thus, the overall C-index ($C_O$) can be split up
into a between-cluster C-index ($C_B$) and a within-cluster C-index ($C_W$).
Let $i=1,\ldots,n$ define the cluster and let $ij$ be the subset $j$ of the
cluster $i$ ($j=1,\ldots,n_{i}$). We also denote by $ij$ and $ij^{\prime}$
two patients from the same cluster $i$ and by $ij$ and $i^{\prime}j^{\prime}$
two patients from two different clusters $i$ and $i^{\prime}$ $(i \neq
i^{\prime})$. For simplicity, we consider no ties, even if it can handle
similarly to the case in Section 2.2.2 when there are ties. Then the
estimated within-cluster C-index ($\hat{C}_W$) is given by
\begin{equation*}
\widehat C_{W} = \frac{1}{n} \sum_{i=1}^{n} \biggl [ \dfrac{
\sum_{j=1}^{n_{i}} \sum_{j^{\prime}=1}^{n_{i}} \delta_{ij}
I(y_{ij}<y_{ij^{\prime}}) I\left(\widehat \eta^{(m)}_{ij} >\widehat
\eta^{(m)}_{ij^{\prime}} \right) } {\sum_{j=1}^{n_{i}}
\sum_{j^{\prime}=1}^{n_{i}} \left\{\delta_{ij} I(y_{ij}<y_{ij^{\prime}})
\right\}} \biggl ],
\end{equation*}
where $\widehat \eta^{(m)}_{ij}=NN(\bx_{ij}; \boldsymbol{%
\widehat{w}}, \boldsymbol{\widehat\beta})$ and the frailty terms are not
included directly in the calculation of the within-cluster concordance since
they are the same for the compared patients in each pair. Next, the
estimated between-cluster C-index ($\hat{C}_B$) considers only comparison
between patients of different clusters and includes the estimated frailty
terms; it is given by
\begin{equation*}
\widehat C_{B} = \dfrac{ \sum_{i=1}^{n} \sum_{j=1}^{n_{i}} \biggl[ %
\sum_{i^{\prime}=1}^{n} \sum_{j^{\prime}=1}^{n_{i}^{\prime}} \delta_{ij}
I(y_{ij}<y_{i^{\prime}j^{\prime}}) I\left(\widehat \eta_{ij} >\widehat
\eta_{i^{\prime}j^{\prime}} \right) \biggl] } { \sum_{i=1}^{n}
\sum_{j=1}^{n_{i}} \biggl[ \sum_{i^{\prime}=1}^{n}
\sum_{j^{\prime}=1}^{n_{i}^{\prime}} \left\{\delta_{ij}
I(y_{ij}<y_{i^{\prime}j^{\prime}}) \right\} \biggl ] },
\end{equation*}
where $\widehat \eta_{ij}=\widehat \eta^{(m)}_{ij} + \widehat v_{i} =NN(\bx_{ij}; \boldsymbol{\widehat{w}}, \boldsymbol{\widehat\beta})
+\widehat v_{i}$ and $\widehat v_{i}=\log \widehat u_{i}$. Thus, the
estimated overall C-index ($\hat{C}_O$) can be expressed as a weighted mean
of $\hat{C}_B$ and $\hat{C}_W $ (Oirbeek and Lesaffre, 2010), given by
\begin{equation}
\widehat C_{O} = \frac{n_{T,comp}}{n_{W,comp}} \widehat C_{W} + \frac{%
n_{T,comp}}{n_{B,comp}} \widehat C_{B},
\end{equation}
where $n_{T,comp}$ is the number of comparable pairs, and $n_{W,comp}$ and $%
n_{B,comp}$ are the number of comparable within-and between-cluster pairs,
respectively. Note that $\widehat C_{B}$ can be easily calculated based on
the function, concordance-index(), in the \texttt{lifelines} of Python.

\subsection*{A.3 Online learning for the DNN-FM}

Since the loss function of the DNN-Cox model does not naturally decouple,
it causes computational difficulties in large data sets. To overcome this
difficulty, Tarkhan and Simon (2022) proposed an online framework.
In this section, we extend the online framework to DNN-FM
by a simple modification (14) of $h_{p}$ in (10).

Let $D_{s}$ be a set of random samples of size $s_{i}\geq 0$ drawn from the
population of each patient (or cluster) $i=1,...,n$, where $s_{i}(\leq n_{i})
$ are non-negative integers. Under the assumption of no ties and no
censoring, define the profiled h-likelihood from the mini-batch $D_{s}$ as
\begin{equation}
h_{p}^{(s)}(\boldsymbol{\theta},\mathbf{v})=\sum_{i:s_{i}>0}
\sum_{j=1}^{s_{i}}\left[ \eta _{ij}-\log \left\{ \sum_{(k,l)\in
R_{ij}^{(s)}}\exp (\eta _{kl})\right\} +\frac{v_{i}-\exp (v_{i})}{
n_{i}\alpha }+c_{i}(\alpha ,n_{i})\right] ,  \label{eq:minibatch}
\end{equation}
where $c_{i}(\alpha ,n_{i})=\{-\alpha ^{-1}\log \alpha -\log \Gamma (\alpha
^{-1})-a_{i}(\alpha ,n_{i})\}/n_{i}$ and $R_{ij}^{(s)}=\{(k,l):y_{kl}\geq
y_{ij}~\mathrm{and}~(i,j,k,l)\in D_{s}\}$ is the risk set at the $(i,j)$th
ordered failure time $y_{ij}$. Note here that profiled h-likelihood (14) from
the mini-batch gives $h_{p}^{(s)}(\boldsymbol{\theta},\mathbf{v})=h_{p}(%
\boldsymbol{\theta},\mathbf{v})$ when $D_{s}=D_{n}$. Let $U_{%
\boldsymbol{\beta}}^{(s)}(\boldsymbol{\theta},\mathbf{v})$, $U_{\alpha
}^{(s)}(\boldsymbol{\theta},\mathbf{v})$, and $U_{\mathbf{v}}^{(s)}(%
\boldsymbol{\theta},\mathbf{v})$ be the score functions of profiled
h-likelihood from $D_{s}$ with respect to $\boldsymbol{\beta}$, $\alpha $,
and $\mathbf{v}$, respectively,
\begin{equation*}
U_{\boldsymbol{\beta}}^{(s)}(\boldsymbol{\theta},\mathbf{v})=\frac{\partial
h_{p}^{(s)}(\boldsymbol{\theta},\mathbf{v})}{\partial \boldsymbol{\beta}}%
,\qquad U_{\alpha }^{(s)}(\boldsymbol{\theta},\mathbf{v})=\frac{\partial
h_{p}^{(s)}(\boldsymbol{\theta},\mathbf{v})}{\partial \alpha },\quad \text{%
and}\quad U_{\mathbf{v}}^{(s)}(\boldsymbol{\theta},\mathbf{v})=\frac{%
\partial h_{p}^{(s)}(\boldsymbol{\theta},\mathbf{v})}{\partial \mathbf{v}}.
\end{equation*}
Then, if $s_{i}>0$ for some $i$ and $s_{j}=0$ for all $j\neq i$, we have the
following Theorem \ref{thm1}.

\begin{theorem}
\label{thm1}
Let $\boldsymbol{\theta}^{\ast }=(\boldsymbol{\beta}^{\ast
},\alpha ^{\ast })$ be the vector of true values of fixed parameters and $%
\widetilde{\mathbf{v}}$ be the mode of profiled h-likelihood at $\theta
=\theta ^{\ast }$, then
\begin{equation*}
E\left[ U_{\boldsymbol{\beta}}^{(s)}(\boldsymbol{\theta}^{\ast },\widetilde{%
\mathbf{v}})\right] =0\quad \text{and}\quad E\left[ U_{\alpha }^{(s)}(%
\boldsymbol{\theta}^{\ast },\widetilde{\mathbf{v}})\right] =0.
\end{equation*}
\end{theorem}

\noindent \textbf{Remark 1:} Tarkhan and Simon (2022) studied the online
learning framework for the DNN-Cox model. Theorem \ref{thm1} extends the
framework to the DNN-FM, with restriction that a mini-batch should be
sampled within a cluster.

\begin{theorem}
\label{thm2} Let $\mathbf{v}^{\ast }$ be the vector of realized values of
random parameters (i.e. log-frailties), then
\begin{equation*}
E\left[ U_{\mathbf{v}}^{(s)}(\boldsymbol{\theta}^{\ast },\mathbf{v}^{\ast })%
\right] \rightarrow 0\quad \text{as }n_{i}\rightarrow \infty ~\mathrm{for~all%
}~i.
\end{equation*}
\end{theorem}

\noindent \textbf{Remark 2:} When the cluster size $n_{i}$ approaches
infinity for all $i$, Theorem \ref{thm2} shows that the frailty predictors
converge in probability to their true realized values. It implies that the
frailty predictors approach the fixed effect estimators of $\mathbf{v}$ of
the Cox model with fixed parameters $\mathbf{v}$. Therefore, the online
learning framework of Tarkhan and Simon (2022) can be directly used for
DNN-FM when $n_{i}\rightarrow \infty $. In this case, mini-batches can be
drawn from multiple clusters.

\subsection*{A.3.1 Proof of Theorem 1}
\noindent (a)
Here, it is enough to consider
\begin{equation*}
h_1^{(s)}(\btheta, \bv)
= \sum_{i:s_i>0} \sum_{j=1}^{s_i} \left[
	 \eta_{ij} - \log \left\{ \sum_{(k,l) \in R_{ij}^{(s)}} \exp(\eta_{kl}) \right\}
	 % + \frac{v_i - \exp(v_i)}{n_i \alpha} + c_i(\alpha, n_i)
\right],
\end{equation*}
since $\bbeta$ does not involve the remaining terms of profiled h-likelihood.
Here, $\eta_{ij} = \eta_{ij}^{(m)} + v_i$ and
$ \eta_{ij}^{(m)}=\eta^{(m)}(\bx_{ij}; \bbeta)=NN(\bx_{ij}; \boldsymbol{w}, \boldsymbol{\beta})$.
Analogous to Tarkhan and Simon (2022),
we define a counting process $dN_{ij}(t)$ as
$$
\int_a^b g(t) dN_{ij}(t) = g(t_{ij}) I(t_{ij} \in [a, b]),
$$
and define $dN^{(s)}(t) = \sum_{i:s_i>0} \sum_{j=1}^{s_i} dN_{ij}(t)$
to be a counting process for failure times over all patients in $D_s$
under the assumption that the failure time process is absolutely continuous
with respect to Lebsegue measure on time,
which implies that there is no ties at any time $t$.
Then, $h_1^{(s)}(\btheta, \bv) $ can be expressed as
$$
h_1^{(s)}(\btheta, \bv)
= \sum_{i:s_i>0} \sum_{j=1}^{s_i} \eta_{ij}
- \sum_{i:s_i>0} \sum_{j=1}^{s_i}
\int_{0}^{\tau}
\log \left\{ \sum_{(k,l) \in R_{ij}^{(s)}}
M_{kl}(t) \exp(\eta_{kl}) \right\} dN_{ij}(t),
$$
where $\tau$ is the duration of the study,
and its derivative is
\begin{align*}
U_{\beta}^{(s)}(\btheta, \bv)=\frac{\partial h_1^{(s)}(\btheta, \bv)}{\partial \bbeta}
&= \sum_{i:s_i>0} \sum_{j=1}^{s_i} \eta'(\bx_{ij}; \bbeta)
- \sum_{i:s_i>0} \sum_{j=1}^{s_i} \int_{0}^{\tau}
\sum_{k, l} w_{kl}(\btheta, \bv) \eta'(\bx_{kl}; \bbeta) dN_{ij}(t)
\\
&= \sum_{i:s_i>0} \sum_{j=1}^{s_i} \eta'(\bx_{ij}; \bbeta)
- \sum_{i:s_i>0} \sum_{j=1}^{s_i} \int_{0}^{\tau}
w_{ij}(\btheta, \bv) \eta'(\bx_{ij}; \bbeta) dN^{s}(t),
\end{align*}
where $\eta'(\bx_{ij}; \bbeta)$ is the gradient of $\eta^{(m)}(\bx_{ij}; \bbeta)$ with respect to $\beta$,
$w_{ij}(\btheta, \bv) = \frac{M_{ij}(t) \exp\{\eta^{(m)}(\bx_{ij};\bbeta)+v_i\}}
{\sum_{k,l}M_{kl}(t) \exp\{\eta^{(m)}(\bx_{kl};\bbeta)+v_i\}}
$
is a weight proportional to the hazard of failure
and $M_{ij}(t)$ is an indicator representing
whether $ij$ is at risk at time $t$, i.e., $t_{ij} \geq t$.
Thus, the score function $U_{\beta}^{(s)}(\btheta^*, \bvtil)$ is given by
$$
U_{\beta}^{(s)}(\btheta^*, \bvtil)
= \sum_{i:s_i>0} \sum_{j=1}^{s_i} \eta'(\bx_{ij}; \bbeta^*)
- \sum_{i:s_i>0} \sum_{j=1}^{s_i} \int_{0}^{\tau}
w_{ij}(\bbeta^*, \bvtil) \eta'(\bx_{ij}; \bbeta^*) dN^{s}(t),
$$
and it is enough to show that
$E(U_{\beta}^{(s)}(\btheta^*, \bvtil)) = E(E(U_{\beta}^{(s)}(\btheta^*, \bvtil) | \bv=\bv^*))=0$.
As in Tarkhan and Simon (2022), we have
$$
E\left(\sum_{i:s_i>0} \sum_{j=1}^{s_i} \eta'(\bx_{ij}; \bbeta^*) \Big| \bv=\bv^* \right)
= \sum_{i:s_i>0} \sum_{j=1}^{s_i} \int_{0}^{\tau}
E \left( w_{ij}(\bbeta^*, \bv^*) \eta'(\bx_{ij}; \bbeta^*) dN^{s}(t) \right),
$$
Then the score function becomes
\begin{align*}
& E(U_{\beta}^{(s)}(\btheta^*, \bvtil) |\bv=\bv^*)
= E\left(
\sum_{i:s_i>0} \sum_{j=1}^{s_i} \eta'(\bx_{ij}; \bbeta^*)
- \sum_{i:s_i>0} \sum_{j=1}^{s_i} \int_{0}^{\tau}
w_{ij}(\bbeta^*, \bvtil) \eta'(\bx_{ij}; \bbeta^*) dN^{s}(t)
\right)
\\
&= \sum_{i:s_i>0} \sum_{j=1}^{s_i} \int_{0}^{\tau}
E \left[ w_{ij}(\bbeta^*, \bv^*) \eta'(\bx_{ij}; \bbeta^*) dN^{s}(t) \right]
- \sum_{i:s_i>0} \sum_{j=1}^{s_i} \int_{0}^{\tau}
E \left[ w_{ij}(\bbeta^*, \bvtil) \eta'(\bx_{ij}; \bbeta^*) dN^{s}(t) | \bv=\bv^* \right]
\\
&= \sum_{i:s_i>0} \sum_{j=1}^{s_i} \int_{0}^{\tau}
\eta'(\bx_{ij}; \bbeta^*)
E \left[ \{w_{ij}(\bbeta^*, \bv^*) -  w_{ij}(\bbeta^*, \bvtil)\}
dN^{s}(t) | \bv=\bv^* \right].
\end{align*}
If the mini-batch is sampled within the $i$-th cluster only,
\begin{align*}
w_{ij}(\bbeta^*, \bv^*) -  w_{ij}(\bbeta^*, \bvtil)
&= \frac{M_{ij}(t) \exp\{\eta^{(m)}(\bx_{ij};\bbeta^*)+v^*_i\}}
{\sum_{l=1}^{s_i} M_{il}(t) \exp\{\eta^{(m)}(\bx_{il};\bbeta^*)+v^*_i\}}
- \frac{M_{ij}(t) \exp\{\eta^{(m)}(\bx_{ij};\bbeta^*)+\vtil_i\}}
{\sum_{l=1}^{s_i} M_{il}(t) \exp\{\eta^{(m)}(\bx_{il};\bbeta^*)+\vtil_i\}}
\\
&= \frac{M_{ij}(t) \exp\{\eta^{(m)}(\bx_{ij};\bbeta^*)\}}
{\sum_{l=1}^{s_i} M_{il}(t) \exp\{\eta^{(m)}(\bx_{il};\bbeta^*)\}}
- \frac{M_{ij}(t) \exp\{\eta^{(m)}(\bx_{ij};\bbeta^*)\}}
{\sum_{l=1}^{s_i} M_{il}(t) \exp\{\eta^{(m)}(\bx_{il};\bbeta^*)\}}
\\
&=0,
\end{align*}
which leads to $E(U_{\beta}^{(s)}(\btheta^*, \bvtil)) = 0$.

\noindent (b)
The score function with respect to $\alpha$ is
$$
U_{\alpha}^{(s)}(\btheta, \bv)
= \frac{\partial h_p^{(s)}(\btheta, \bv)}{\partial \alpha}
= \frac{s_i}{n_i}  \frac{\partial}{\partial \alpha}
\left[
\frac{ \log u_i - u_i }{\alpha}
%\frac{\log \util_i - \util}{\alpha}
- \frac{\log \alpha}{\alpha}
- \log \Gamma \left( \alpha^{-1} \right)
- a_i(\alpha, n_i)
\right]
$$
where
$
a_i(\alpha, n_i) =
\left(n_i+\alpha^{-1}\right) \left\{ \log \left(n_i+\alpha^{-1}\right) - 1 \right\}
- \log \Gamma \left(n_i+\alpha^{-1}\right).
$
Since
\begin{equation*}
u_i | \mathbf{y}_i^{*} \sim \text{Gamma} \left( \delta_{i+}+\alpha^{-1},
(\Lambda_{i+}+\alpha^{-1})^{-1} \right)
\end{equation*}
and $\util_i=\util_i(\alpha)=(n_{i}+ \alpha^{-1})/(\Lambda_{i+} +\alpha^{-1})=E(u_{i}| \mathbf{y}_i^{*})$,
\begin{align*}
U_{\alpha}^{(s)}(\btheta^*, \bvtil)
&= U_{\alpha}^{(s)}(\btheta, \bv)|_{\btheta=\btheta^*, \bv= \bvtil}
\\
& = \frac{s_i}{n_i} \frac{1}{\alpha^{2}}
\left[
 (u_i - \log u_i - 1)
+\log \left(n_i+\frac{1}{\alpha}\right) - \log \left(\frac{1}{\alpha}\right)
-\psi \left(n_i+\frac{1}{\alpha}\right) + \psi \left(\frac{1}{\alpha}\right)
\right]\Big|_{\btheta=\btheta^*, \bv= \bvtil}
\\
& = \frac{s_i}{n_i} \frac{1}{\alpha^{*2}}
\left[
 (\util_i - \log \util_i - 1)
+\log \left(n_i+\frac{1}{\alpha^{*}}\right) - \log \left(\frac{1}{\alpha^{*}}\right)
-\psi \left(n_i+\frac{1}{\alpha^{*}}\right) + \psi \left(\frac{1}{\alpha^{*}}\right)
\right]
\\
& = \frac{s_i}{n_i} \frac{1}{\alpha^{*2}}
\left[
 (\util_i - 1)
+ \log \left( \Lambda_{i+} + \frac{1}{\alpha^*} \right)
- \log \left(\frac{1}{\alpha^{*}}\right)
-\psi \left(n_i+\frac{1}{\alpha^{*}}\right) + \psi \left(\frac{1}{\alpha^{*}}\right)
\right]
\\
& = \frac{s_i}{n_i} \frac{1}{\alpha^{*2}}
\left[
 E(u_i|\by^*) - 1 - E(\log u_i |\by^*)
- \log \left(\frac{1}{\alpha^{*}}\right)
+ \psi \left(\frac{1}{\alpha^{*}}\right)
\right]
\end{align*}
where $\psi(\cdot)$ is the digamma function.
Thus, we have
$$
E \left[ U_{\alpha}^{(s)}(\btheta^*, \bvtil) \right]
= \frac{s_i}{n_i} \frac{1}{\alpha^{*2}}
\left[
 E(u_i) - 1 - E(\log u_i)
- \log \left(\frac{1}{\alpha^{*}}\right)
+ \psi \left(\frac{1}{\alpha^{*}}\right)
\right]
= 0
$$
since $E(u_{i})=1$ and $E(\log u_{i})= \psi(1/\alpha^*) +\log(\alpha^*)$.

\subsection*{A.3.2  Proof of Theorem 2}

Let $\bz_{ij}$ be the one-hot encoded vector of cluster number,
so that $\bz_{ij}^T \bv = v_i$,
then the predictor $\eta_{ij}$ can be expressed as
$\eta_{ij} = \eta^{(m)}(\bx_{ij}; \bbeta) + v_i = \eta^*(\bx_{ij}, \bz_{ij}; \bbeta, \bv)$.
For example,
$$
\eta_{ij} = \bx_{ij}^T \bbeta + v_i = (\bx_{ij}^T, \bz_{ij}^T) (\btheta, \bv).
$$
Define $h_1^{(s)}(\btheta, \bv)$ as
$$
h_1^{(s)}(\btheta, \bv)
= \sum_{i:s_i>0} \sum_{j=1}^{s_i} \left[ \eta_{ij}
- \log \left\{ \sum_{(k,l) \in R_{ij}^{(s)}} \exp(\eta_{kl}) \right\} \right],
$$
then the profiled h-likelihood in \eqref{eq:minibatch} can be expressed as
$$
h_p^{(s)}(\btheta, \bv)
= h_1^{(s)}(\btheta, \bv)
+ \sum_{i:s_i>0} \sum_{j=1}^{s_i}
\left[ \frac{v_i - \exp(v_i)}{n_i \alpha} + c_i(\alpha, n_i) \right].
$$
Thus, $h_1^{(s)}(\btheta, \bv)$ is equivalent to the log-partial likelihood (Tarkhan and Simon, 2022)
when $\bv$ is treated as the fixed parameters,
and the remaining terms does not depend on $\bbeta$.
Therefore, by the results of Tarkhan and Simon (2022),
$$
E\left[U_{\bbeta}^{(s)}(\btheta^*, \bv^*)\right] = 0,
$$
and
$$
E\left[U_{\bv}^{(s)}(\btheta^*, \bv^*)\right]
= E \left[
\sum_{i:s_i>0} \sum_{j=1}^{s_i} \frac{\partial}{\partial \bv}
\left[ \frac{v_i - \exp(v_i)}{n_i \alpha} + c_i(\alpha, n_i) \right] \right]
_{\btheta=\btheta^*, \bv=\bv^*}.
$$
When $n_i \to \infty$,
$$
\frac{\partial}{\partial v_i}
\left[ \frac{v_i - \exp(v_i)}{n_i \alpha^*}
+ c_i(\alpha^*, n_i) \right]
= \frac{1}{n_i} \left[ \frac{1 - \exp(v_i^*)}{\alpha^*} \right] \to 0.
$$
Thus,
$$
E\left[U_{\bv}^{(s)}(\btheta^*, \bv^*)\right] \to 0.
$$

\section*{Reference}

\begin{description}
\item  Balan, T. A. and Putter, H. (2019). frailtyEM: An R package for estimating
semiparametric shared frailty models. {\it Journal of Statistical Software}, {\bf 90}, 1--29.

\item  Breslow, N.E. (1972) Discussion on Professor Cox's paper. \textit{%
Journal of the Royal Statistical Society B} 34, 216--217.

\item  Chee, C.-S., Ha, I.D., Seo, B. and Lee, Y. (2021). Semiparametric
estimation for nonparametric frailty models using nonparametric maximum
likelihood approach. \textit{Statistical Methods in Medical Research},
\textbf{30}, 2485--2502.

\item  Fan, J., Ma, C. and Zhong, Y. (2021). A selective overview of deep
learning. \textit{Statistical Science}, 36, 264--290.

\item  Farrell, M. H., Liang, T. and Misra, S. (2021). Deep neural networks
for estimation and inference. \textit{Econometrica}, 89, 181--213

\item  Goodfellow, I., Bengio, Y., and Courville, A. (2016). \textit{Deep
learning}. MIT Press.

\item  Gorfine, M. and Zuker, D.M. (2023). Shared frailty models for complex
survival data: a review of recent advances. \textit{Annual Review of
Statistics and Its Application}, 10, 1-23.

\item  Graf, E., Schmoor, C., Sauerbrei, W. and Schumacher, M. (1999)
Assessment and comparison of prognostic classification schemes for survival
data. \textit{Statistics in Medicine}, 18, 2529-2545.

\item  Gu, M. G., Sun, L. and Huang, C. (2004). A universal procedure for
parametric frailty models. \textit{J. Statist. Comput. Simulation}, 74, 1--13.

\item  Tarkhan,A. and Simon, N. (2022). An online framework for survival
analysis: reframing Cox proportional hazards model for large data sets and
neural networks. \textit{Biostatistics}, in press.

\item  Ha, I. D., Jeong, J. H. and Lee, Y. (2017). \textit{Statistical
modeling of survival data with random effects: h-likelihood approach}.
Springer.

\item  Ha, I.D., Lee, Y. and Song, J.K. (2001) Hierarchical likelihood
approach for frailty models. \textit{Biometrika}, 88, 233-243.

\item  Harrell, F.E, Lee, K.L. and Mark, D.B. (1996) Multivariable
prognostic models: issues in developing models, evaluating assumptions and
adequacy, and measuring and reducing errors. \textit{Statistics in Medicine}%
, 15, 361--387.

\item  Hsu, L., Gorfine, M., Malone, K. (2007). Effect of frailty
distribution misspecification on marginal regression estimates and hazard
functions in multivariate survival analysis. \textit{Statistics in Medicine}, {\bf 26}, 4657--4678.

\item  Jeon, J., Hsu, L. and Gorfine, M. (2012). Bias correction in the
hierarchical likelihood approach to the analysis of multivariate survival
data. \textit{Biostatistics} 13, 384--397.

\item  Kvamme, H., Borgan and Scheel, I. (2019). Time-to-event prediction
with neural networks and Cox Regression. \textit{Journal of Machine Learning
Research}, 20, 1-30.

\item  LeCun, Y., Bengio, Y., and Hinton, G. (2015). Deep learning. \textit{%
Nature} 521:436.

\item  Lee, Y. and Nelder, J. A. (1996). Hierarchical generalized linear
models (with discussion). \textit{Journal of the Royal Statistical Society,
Series B}, \textbf{58}, 619--678.

\item  Lee, Y., Nelder, J. A., and Pawitan, Y. (2017). \textit{Generalised
linear models with random effects: Unified analysis via h-likelihood}. 2nd
edn., Chapman and Hall: London.

\item  Mandel, F., Ghosh, R. P. and Barnett, I. (2022) Neural networks for
clustered and longitudinal data using mixed effects models. \textit{%
Biometrics}, in press.

\item  Montesinos-Lopez, et al. (2021). Application of a Poisson deep neural
network model for the prediction of count data in genome-based prediction.
\textit{Plant Genome}. e20118.

\item  Park, E. and Ha, I.D. (2019) Penalized variable selection for
accelerated failure time models with random effects. \textit{Statistics in
Medicine}, 38, 878--892.

\item  Polson, N.G. and Sokolov, V. (2017). Deep learning: a Bayesian
perspective. \textit{Bayesian Analysis}, 12, 1275-1304.

\item  Rodrigo, H. and Tsokos, C. (2020) Bayesian modelling of nonlinear
poisson regression with artificial neural networks. \textit{Journal of
Applied Statistics}, 47, 757--774.

\item  Searle, S. R., Casella, G., and McCulloch, C. E. (1992). \textit{%
Variance components}. Wiley: New York.

\item  Sun, T., Wei, Y., Chen,W., Ding, Y. (2020). Genome-wide association
study-based deep learning for survival prediction. \textit{Statistics in
Medicine}, 39, 4605--4620.

\item  Sylvester R, van der Meijden APM, Oosterlinck W, Witjes J, Bouffioux
C, Denis L, Newling DWW, Kurth K. (2006) Predicting recurrence and
progression in individual patients with stage Ta T1 bladder cancer using
EORTC risk tables: a combined analysis of 2596 patients from seven EORTC
trials. \textit{Eur Urol}, 49, 466-477

\item  Tarkhan, A. and Simon, N. (2022). An online framework for survival
analysis: reframing Cox proportional hazards model for large data sets and
neural networks. \textit{Biostatistics}, in press.

\item  Tran, M.-N., Nguyen, N., Nott,D. and Kohn, R. (2020). Bayesian deep
net GLM and GLMM. \textit{Journal of Computational and Graphical Statistics}%
, 29. 97-113.

\item  Van Oirbeek, R and Lesaffre, E. (2010) An application of Harrell's
c-index to PH frailty models. \textit{Statistics in Medicine}, 29, 3160-3171.

\item  Van Oirbeek, R and Lesaffre, E. (2016) Exploring the clustering
effect of the frailty survival model by means of the Brier score. \textit{%
Communications in Statistics-Simulation and Computation}, 45, 3294-3306.
\end{description}

\pagebreak

\begin{figure}[]
\centering
\includegraphics[width=13cm,height=10cm]{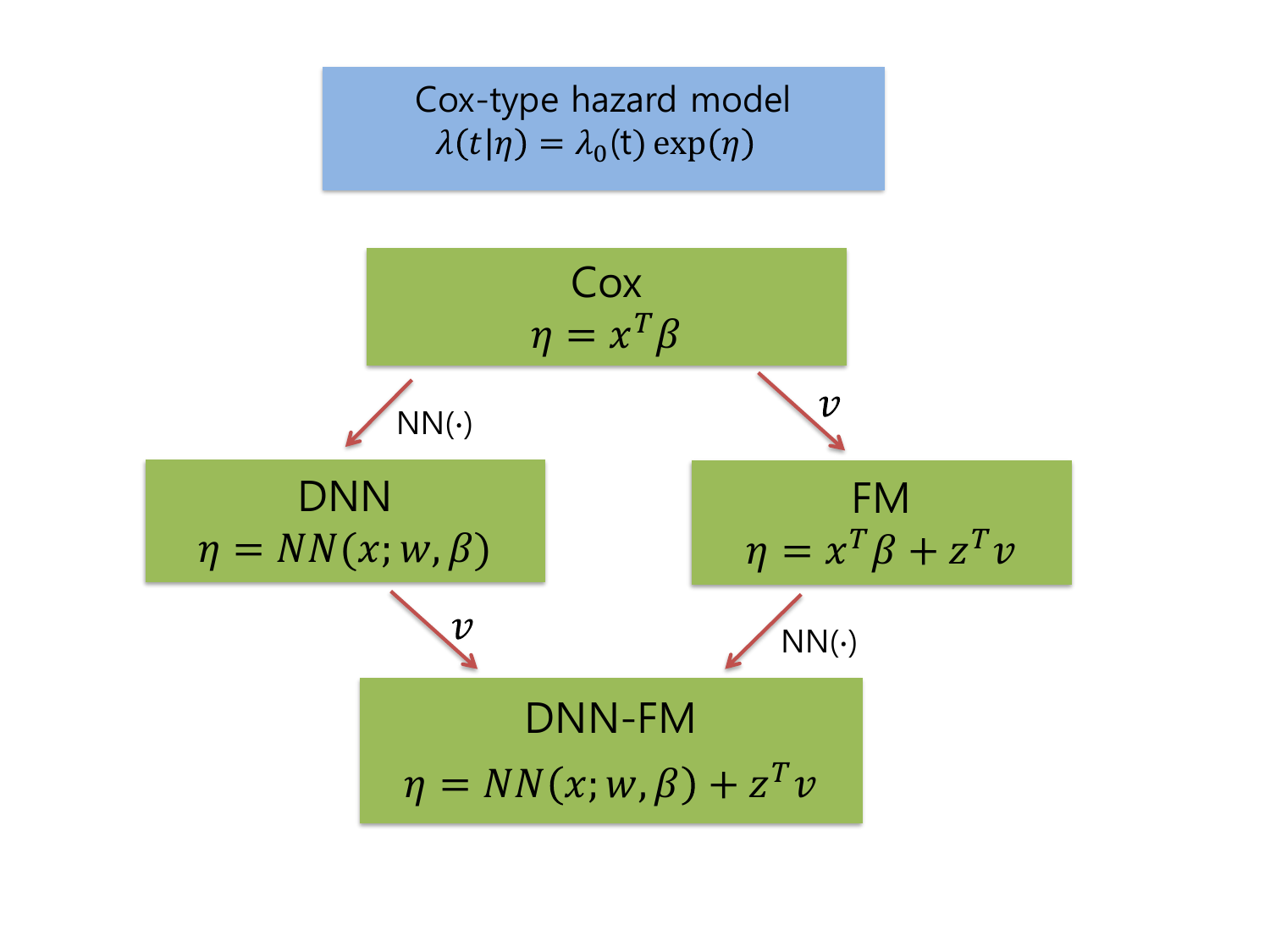}
\par
\vspace{1cm}
\caption{{\protect\small Extension of the Cox model to DNN-FM:~ NN$(\cdot)$,
NN predictor; $v$, log-frailty}}
\end{figure}

\pagebreak

\begin{figure}[]
\centering
\includegraphics[width=15cm,height=10cm]{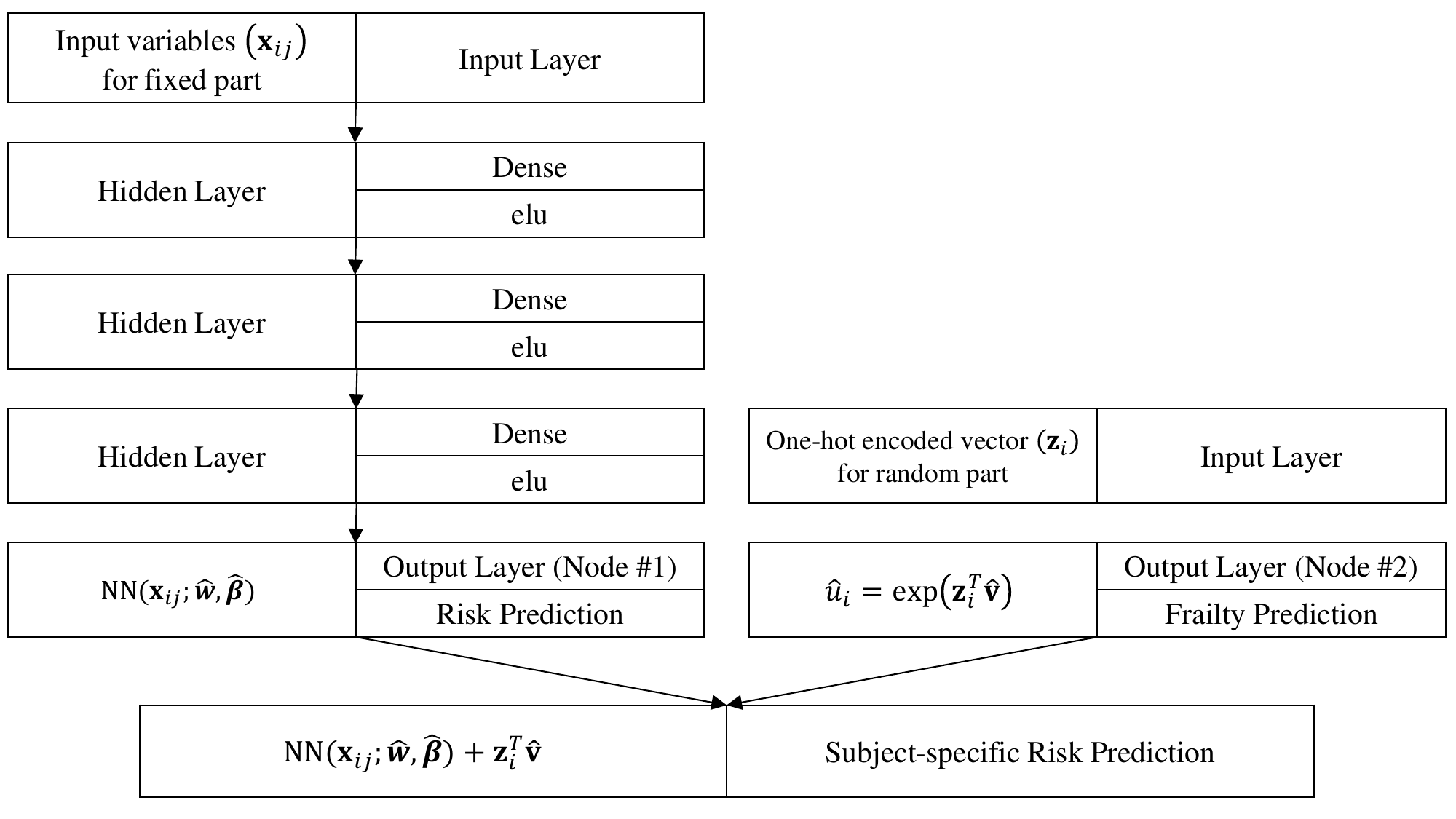}
\par
\vspace{1cm}
\caption{{\protect\small An example of model architecture for DNN-FM}}
\end{figure}

\pagebreak

\begin{figure}[]
\centering
\includegraphics[width=14cm,height=10cm]{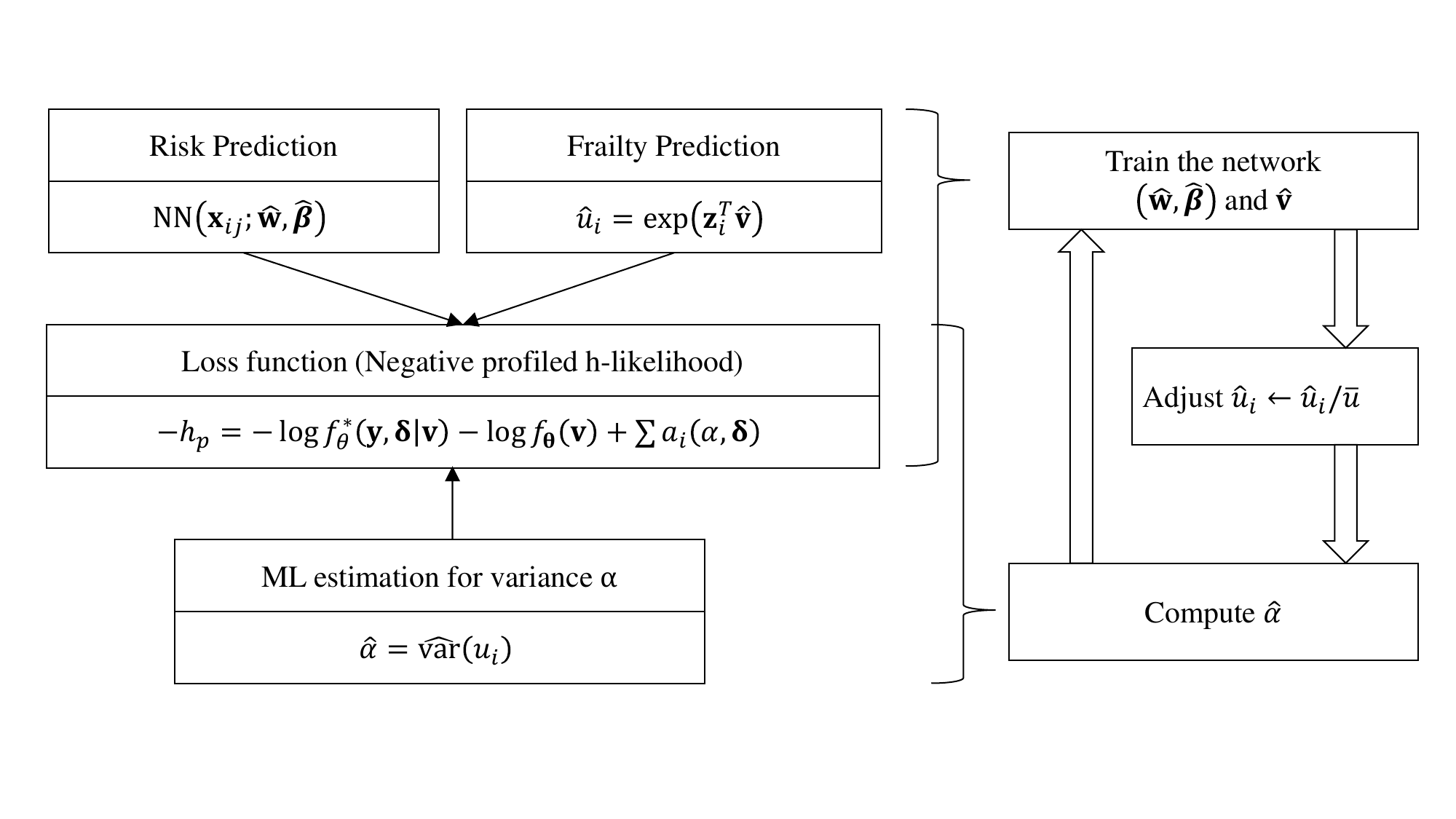}
\par
\vspace{1cm}
\caption{{\protect\small A schematic diagram of DNN-FM fitting procedure; $%
\log f_{\protect\theta}^{*}(\cdot)$, conditional profiled log-likelihood
obtained by profiling out the baseline hazard $\protect\lambda_{0}(t)$. }}
\end{figure}

\pagebreak

\begin{figure}[]
\centering
\includegraphics[width=13cm,height=12cm]{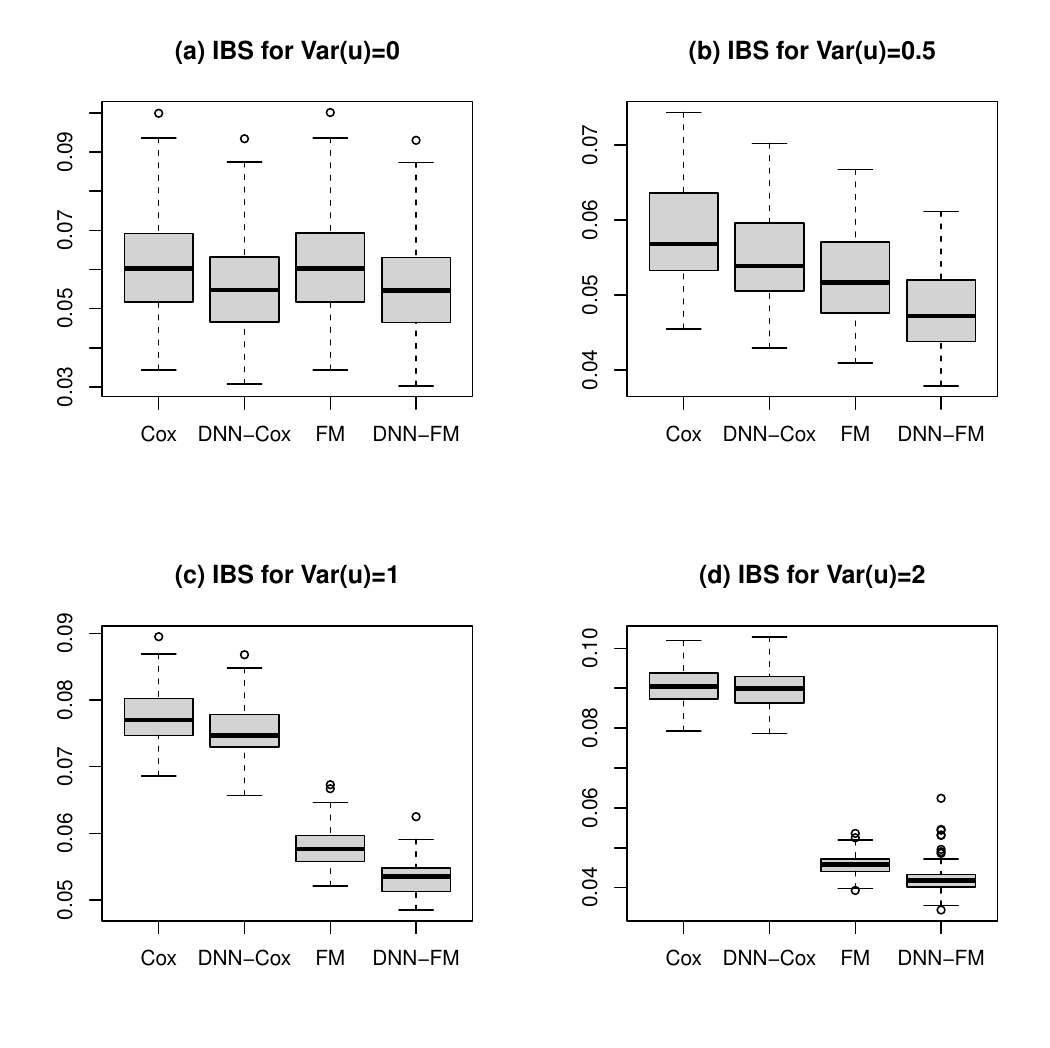}
\par
%\vspace{-3cm}
\caption{{\protect\small 15\% censoring: Box plot of IBS from 100
replications for each frailty variance, var$(u)=\protect\alpha$.}}
\end{figure}

\begin{figure}[]
\centering
\includegraphics[width=13 cm,height=12cm]{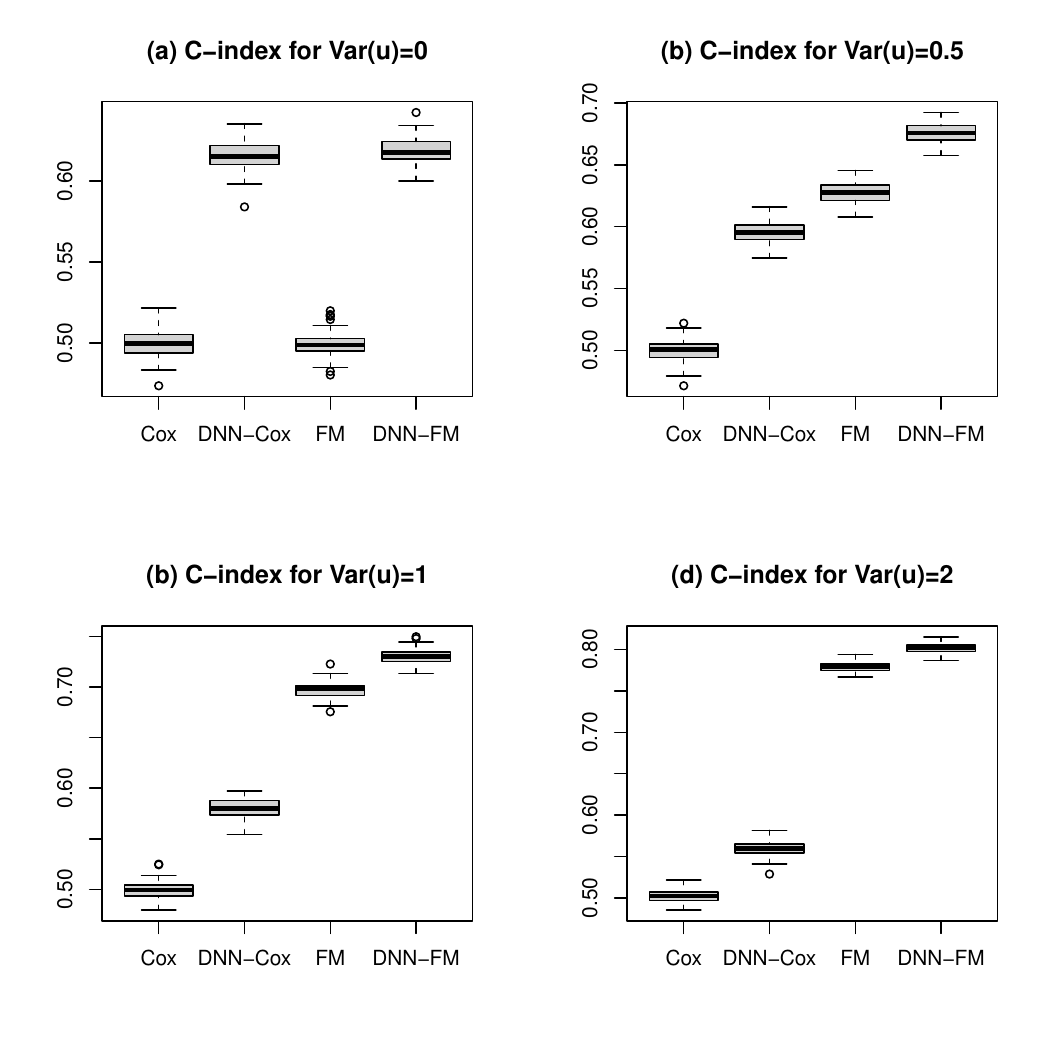}
\par
%\vspace{-3cm}
\caption{{\protect\small 15\% censoring: Box plot of C-index from 100
replications for each frailty variance, var$(u)=\protect\alpha$.}}
\end{figure}

\pagebreak

\begin{figure}[]
\centering
\includegraphics[width=13cm,height=12cm]{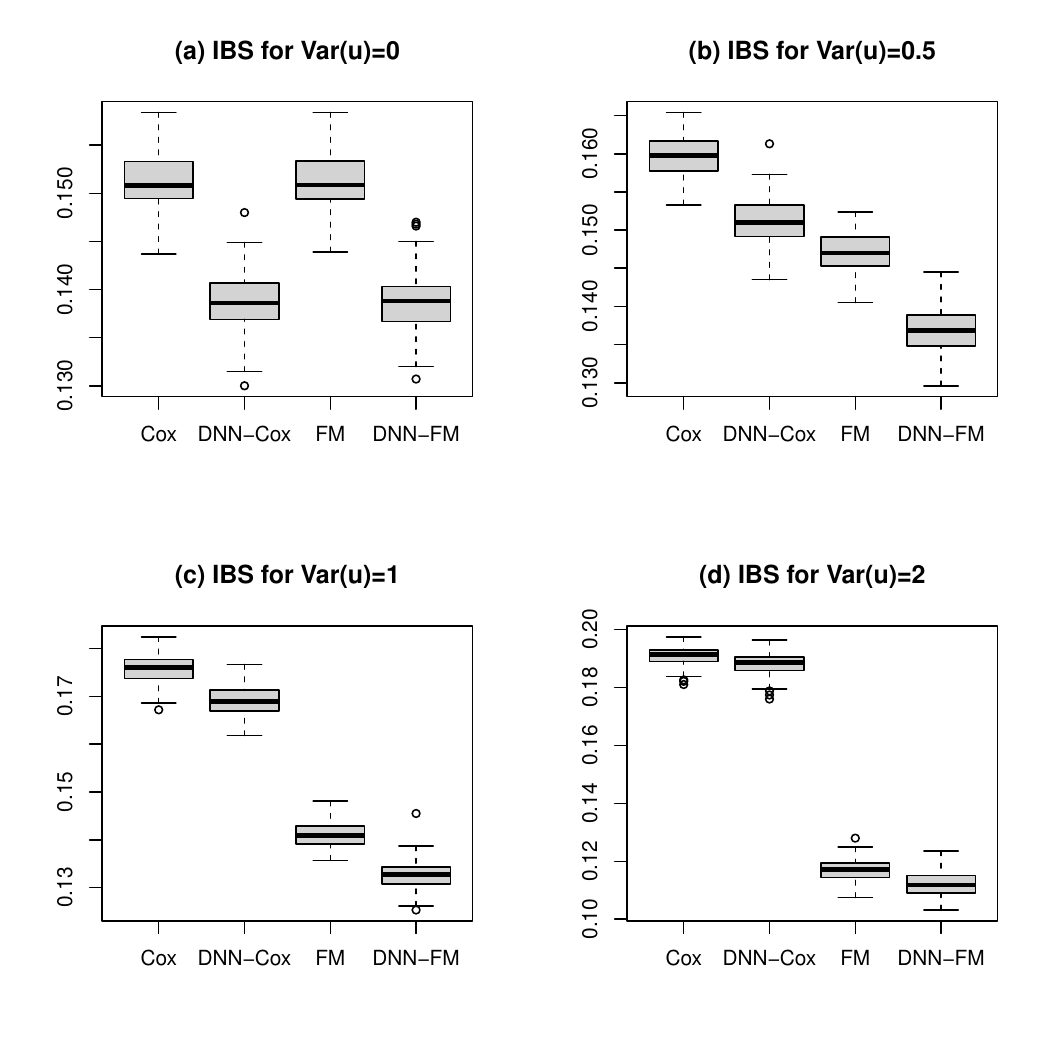}
\par
%\vspace{-3cm}
\caption{{\protect\small 45\% censoring: Box plot of IBS from 100
replications for each frailty variance, var$(u)=\protect\alpha$.}}
\end{figure}

\begin{figure}[]
\centering
\includegraphics[width=13 cm,height=12cm]{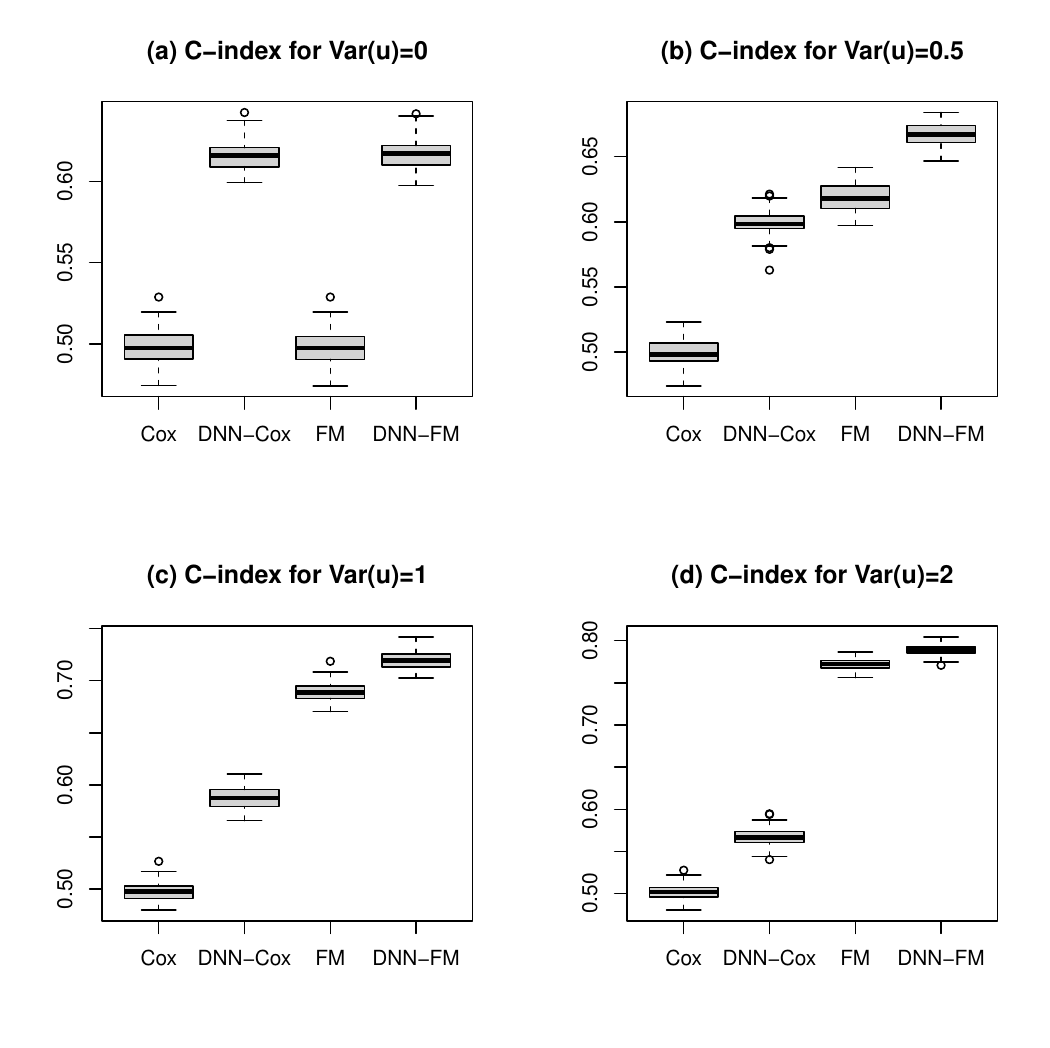}
\par
%\vspace{-3cm}
\caption{{\protect\small 45\% censoring: Box plot of C-index from 100
replications for each frailty variance, var$(u)=\protect\alpha$.}}
\end{figure}

\begin{figure}[]
\centering
\includegraphics[width=15 cm,height=12cm]{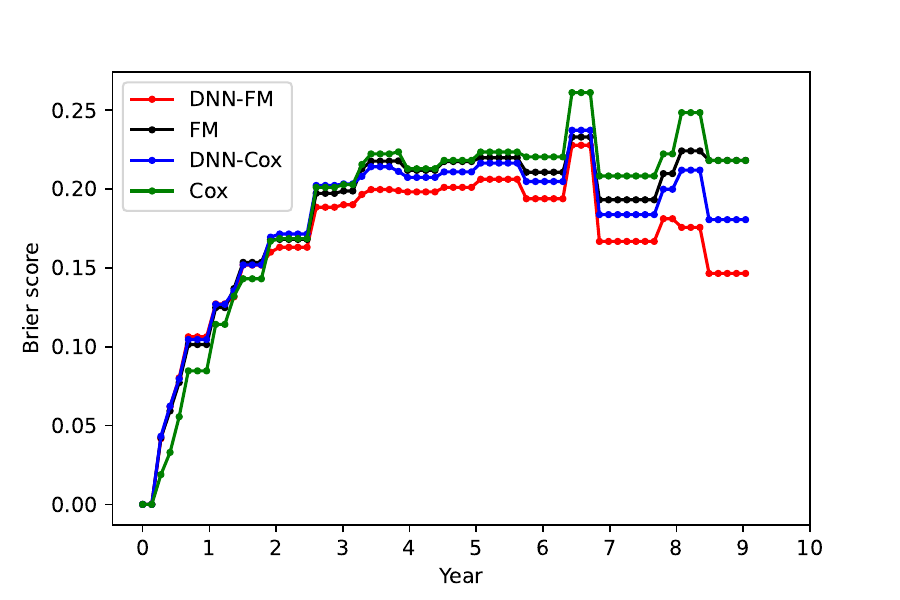}
\par
%\vspace{-3cm}
\caption{{\protect\small Time-dependent Brier score for four survival
prediction models on the test set of the bladder cancer data.}}
\end{figure}

\pagebreak

\begin{table*}[!t]
\caption{Mean (standard deviation) of IBS and C-index from 100 replications
for each frailty variance $\protect\alpha$.}
\label{tb:rmse}\centering
\begin{tabular}{|ll|cccc|}
\hline
&  &  &  &  &  \\
Censoring & Measure & Cox & DNN-Cox & FM & DNN-FM \\
&  &  &  &  &  \\ \hline
15\% & IBS &  &  &  &  \\
& \multirow{2}{*}{$\alpha=0$} & 0.062 & 0.056 & 0.062 & 0.056 \\
&  & (0.0127) & (0.0119) & (0.0127) & (0.0119) \\
& \multirow{2}{*}{$\alpha=0.5$} & 0.058 & 0.055 & 0.053 & 0.048 \\
&  & (0.0066) & (0.0063) & (0.0061) & (0.0055) \\
& \multirow{2}{*}{$\alpha=1$} & 0.077 & 0.075 & 0.058 & 0.053 \\
&  & (0.0043) & (0.0042) & (0.0030) & (0.0026) \\
& \multirow{2}{*}{$\alpha=2$} & 0.090 & 0.090 & 0.046 & 0.042 \\
&  & (0.0047) & (0.0047) & (0.0027) & (0.0042) \\
& C-index &  &  &  &  \\
& \multirow{2}{*}{$\alpha=0$} & 0.499 & 0.615 & 0.499 & 0.618 \\
&  & (0.0085) & (0.0084) & (0.0069) & (0.0078) \\
& \multirow{2}{*}{$\alpha=0.5$} & 0.500 & 0.596 & 0.627 & 0.675 \\
&  & (0.0082) & (0.0087) & (0.0081) & (0.0076) \\
& \multirow{2}{*}{$\alpha=1$} & 0.499 & 0.580 & 0.697 & 0.730 \\
&  & (0.0081) & (0.0093) & (0.0078) & (0.0070) \\
& \multirow{2}{*}{$\alpha=2$} & 0.502 & 0.559 & 0.779 & 0.802 \\
&  & (0.0072) & (0.0088) & (0.0056) & (0.0051) \\
45\% & IBS &  &  &  &  \\
& \multirow{2}{*}{$\alpha=0$} & 0.151 & 0.139 & 0.151 & 0.139 \\
&  & (0.0031) & (0.0030) & (0.0031) & (0.0030) \\
& \multirow{2}{*}{$\alpha=0.5$} & 0.160 & 0.151 & 0.147 & 0.137 \\
&  & (0.0026) & (0.0032) & (0.0025) & (0.0031) \\
& \multirow{2}{*}{$\alpha=1$} & 0.176 & 0.169 & 0.141 & 0.133 \\
&  & (0.0029) & (0.0031) & (0.0029) & (0.0032) \\
& \multirow{2}{*}{$\alpha=2$} & 0.191 & 0.188 & 0.117 & 0.112 \\
&  & (0.0033) & (0.0038) & (0.0039) & (0.0042) \\
& C-index &  &  &  &  \\
& \multirow{2}{*}{$\alpha=0$} & 0.498 & 0.616 & 0.498 & 0.617 \\
&  & (0.0107) & (0.0087) & (0.0106) & (0.0092) \\
& \multirow{2}{*}{$\alpha=0.5$} & 0.500 & 0.599 & 0.619 & 0.667 \\
&  & (0.0108) & (0.0096) & (0.0108) & (0.0092) \\
& \multirow{2}{*}{$\alpha=1$} & 0.498 & 0.587 & 0.689 & 0.720 \\
&  & (0.0085) & (0.0108) & (0.0088) & (0.0081) \\
& \multirow{2}{*}{$\alpha=2$} & 0.502 & 0.567 & 0.772 & 0.789 \\
&  & (0.0089) & (0.0099) & (0.0068) & (0.0064) \\ \hline
\end{tabular}
\end{table*}

\pagebreak

\begin{table*}[!t]
\caption{Mean (standard deviation) of estimated frailty variance ($\hat
\protect\alpha$) from 100 replications.}
\label{tb:rmse}\centering
\begin{tabular}{|l|cc|}
\hline
&  &  \\
Frailty variance $\alpha$ & FM & DNN-FM \\
&  &  \\ \hline
15\% censoring &  &  \\
\multirow{1}{*}{$\alpha=0$} & 0.006 (0.009) & 0.008 (0.010) \\
\multirow{1}{*}{$\alpha=0.5$} & 0.390 (0.035) & 0.485 (0.050) \\
\multirow{1}{*}{$\alpha=1$} & 0.823 (0.051) & 1.000 (0.062) \\
\multirow{1}{*}{$\alpha=2$} & 1.711 (0.084) & 2.043 (0.094) \\
45\% censoring &  &  \\
\multirow{1}{*}{$\alpha=0$} & 0.008 (0.012) & 0.011 (0.014) \\
\multirow{1}{*}{$\alpha=0.5$} & 0.417 (0.047) & 0.496 (0.054) \\
\multirow{1}{*}{$\alpha=1$} & 0.859 (0.060) & 0.995 (0.090) \\
\multirow{1}{*}{$\alpha=2$} & 1.748 (0.102) & 2.065 (0.123) \\ \hline
\end{tabular}
\end{table*}

\pagebreak

\begin{table*}[!t]
\caption{IBS and C-index for four survival prediction models on the test set
of the bladder cancer data}
\label{tb:rmse}\centering
\begin{tabular}{|l|cccc|}
\hline
&  &  &  &  \\
Measure & Cox & DNN-Cox & FM & DNN-FM \\
&  &  &  &  \\ \hline
IBS & 0.189 & 0.183 & 0.187 & 0.168 \\
C-index & 0.675 & 0.682 & 0.668 & 0.693 \\ \hline
\end{tabular}
\end{table*}

\end{document}